\documentclass{article}
\PassOptionsToPackage{numbers, compress}{natbib}

\usepackage[preprint]{neurips_2026}
\usepackage{float}

\usepackage[utf8]{inputenc} 
\usepackage[T1]{fontenc}    
\usepackage{hyperref}       
\usepackage{url}            
\usepackage{booktabs}       
\usepackage{amsfonts}       
\usepackage{nicefrac}       
\usepackage{microtype}      
\usepackage{xcolor}         
\usepackage{graphicx}
\usepackage{amsmath}
\usepackage{amssymb}
\usepackage{amsfonts}
\usepackage{multirow}
\usepackage{makecell}
\usepackage{multirow}
\usepackage{arydshln}
\usepackage{bbm}
\usepackage{xcolor}

\title{SceneGraphVLM: Dynamic Scene Graph Generation from Video with Vision-Language Models}

\author{%
  Vladislav Makarov \\
  MIRAI \\
  \texttt{makvlad2k03@gmail.com} \\
  \And
  Mark Gizetdinov \\
  MIREA \\
  \texttt{ttt56793@gmail.com} \\
  \And
  Dmitry Yudin \\
  MIRAI, AXXX \\
  \texttt{yudin.da@miriai.org} \\
}

\begin{document}

\maketitle

\begin{abstract}
Scene graph generation provides a compact structured representation for visual perception, but accurate and fast graph prediction from images and videos remains challenging. Recent VLM-based methods can generate scene graphs end-to-end as
structured text, yet often produce long outputs with irrelevant objects and relations. We present SceneGraphVLM, a compact method for image and video scene graph generation with small visual language models. SceneGraphVLM serializes
graphs in a token-efficient TOON format and trains the model in two stages: supervised fine-tuning followed by reinforcement learning with hallucination-aware rewards that balance relation coverage and precision while penalizing unsupported objects and relations. For videos, the model can optionally condition each frame on the previously generated graph, providing lightweight short-term context without tracking or post-processing. We evaluate SceneGraphVLM on PSG, PVSG, and Action Genome. With compact VLMs and vLLM-accelerated decoding, SceneGraphVLM achieves a strong quality-speed trade-off, improves precision-oriented SGG metrics while preserving reasonable recall, and generates complete scene graphs with approximately one-second latency. Code and implementation details are available at:
\url{https://github.com/markus0440/SceneGraphVLM.git}.
\end{abstract}

\section{Introduction}
\label{sec:intro}

Scene graph generation (SGG) represents a visual scene as a compact, interpretable, and machine-readable graph, where nodes denote objects and edges encode their relations. Such graphs are useful for downstream reasoning and decision-making in robotics, navigation, and video analysis. Modern benchmarks extend this formulation from static images, such as Panoptic Scene Graph (PSG)~\cite{yang2022psg}, to videos, such as Panoptic Video Scene Graph (PVSG)~\cite{yang2023pvsg} and Action Genome (AG)~\cite{ji2020action}, where neighboring frames provide useful temporal context and raise additional consistency challenges. Despite this progress, generating accurate, compact, and structurally valid scene graphs remains challenging.

Most SGG methods follow multi-stage pipelines~\cite{xu2017imp},~\cite{zellers2018motifs},~\cite{tang2019vctree}: objects are first detected or segmented, relations are then predicted by a separate module, and the output is often refined with dataset-specific heuristics. Although effective, such systems inherit the brittleness of modular designs: early errors propagate, object-relation consistency is not guaranteed, and transfer across datasets requires additional adaptation. In videos, these issues are amplified by the need to model neighboring-frame context.

Recently, VLM-based approaches have expanded SGG beyond conventional pipelines~\cite{li2024pixels,xu2025llavaspacesgg,dutta2025openworld}. In particular, vision-language models can generate scene graphs end-to-end as structured text~\cite{chen2025compile}. However, this formulation introduces new challenges: graph representations can be long and costly to decode, models may violate the required format, and recall-oriented rewards may encourage overly complete graphs with unsupported objects and relations. As a result, useful information is diluted, making the graph less reliable for downstream perception and decision-making.

\begin{figure}[!h]
    \centering
    \includegraphics[width=0.7\linewidth]{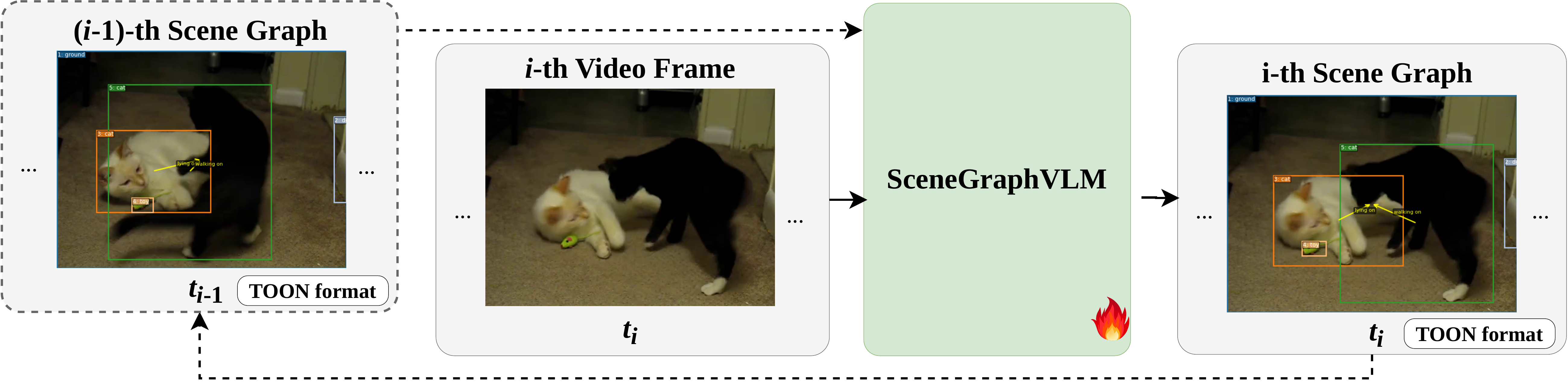}
    \caption{A simplified diagram of the proposed vision-language model-driven SceneGraphVLM method for scene graph generation from video sequences with an optional previous scene graph as input. As a scene graph representation, we use the effective textual TOON format.}
    \label{fig:ga}
\end{figure}

We propose SceneGraphVLM, a compact end-to-end method for image and video scene graph generation with small vision-language models. Unlike multi-stage methods, SceneGraphVLM directly emits a structured textual graph. We represent graphs using the TOON schema~\cite{Schopplich2025TOON}, which preserves explicit graph structure while requiring fewer tokens than conventional formats such as JSON. This shortens responses and accelerates generation without sacrificing machine readability.

Training follows a two-stage procedure: Supervised Fine-Tuning (SFT)~\cite{ouyang2022training} teaches the model to map visual inputs to textual scene graphs, and Group Relative Policy Optimization (GRPO)~\cite{shao2024deepseekmath} further optimizes graph-oriented rewards. Our RL stage builds on R1-SGG~\cite{chen2025compile}, which demonstrated the effectiveness of GRPO for end-to-end SGG. Unlike recall-oriented objectives, our reward also accounts for graph precision by penalizing ungrounded objects, spurious relations, and structural errors, encouraging compact and visually grounded predictions.

For videos, SceneGraphVLM processes frames sequentially and can condition on the graph generated for the previous frame (Figure~\ref{fig:ga}). This provides
a lightweight form of short-term context without requiring a separate tracking
module or complex post-processing. We evaluate this setting separately from
image-only inference because generated previous graphs can also introduce error
propagation.

We instantiate SceneGraphVLM with compact VLMs, including Qwen3.5-0.8B~\cite{qwenteam2026qwen35omnitechnicalreport}, and evaluate it on PSG, PVSG, and Action Genome. Experiments show that a token-efficient representation, two-stage training, and hallucination-aware rewards allow small models to achieve strong quality and speed. With the compact TOON format and vLLM-accelerated inference, SceneGraphVLM generates complete scene graphs in approximately one second, making VLM-based SGG practical for near-real-time scenarios.

Our main contributions are:

\begin{itemize}
\item We propose SceneGraphVLM, a compact end-to-end method for image and video scene graph generation using small vision-language models. It employs a token-efficient TOON representation that preserves objects, bounding boxes, attributes, and relations while reducing generation length compared with JSON.
\item We introduce hallucination-aware GRPO training with graph-oriented rewards that balance relation recall and precision and penalize unsupported objects and relations.
\item We show that SceneGraphVLM achieves a strong quality-speed trade-off on PSG, PVSG, and Action Genome while providing approximately one-second latency for near-real-time inference.
\end{itemize}

\section{Related Work}
\label{sec:rel_works}

\textbf{Scene Graph Generation (SGG).} Scene graph generation represents an image or a video frame as a structured graph whose nodes denote objects and edges denote their relations. Classical SGG methods typically follow multi-stage pipelines: objects are first detected or segmented, and relations are then predicted by a separate module. Although contextual modeling methods such as iterative message passing~\cite{xu2017imp}, Neural Motifs~\cite{zellers2018motifs}, and VCTree~\cite{tang2019vctree} have substantially improved SGG, they still inherit the limitations of modular systems: error propagation, weak object-relation consistency, and costly adaptation to new datasets. In videos, these issues are amplified by frame-to-frame redundancy, object and
relation persistence, and the potential accumulation of errors across neighboring
frames.

\textbf{LLMs for Scene Graph Generation.} With the rise of VLMs, SGG is increasingly formulated as an image-to-text task, where a model directly generates a structured description of objects, bounding boxes, and relations~\cite{li2024pixels},~\cite{xu2025llavaspacesgg},~\cite{dutta2025openworld}. This formulation combines localization and relation prediction in a single generative model and better supports open-vocabulary and prompt-driven settings. We focus on compact open VLMs, primarily Qwen3.5~\cite{qwenteam2026qwen35omnitechnicalreport},
jointly optimizing graph quality, output length, latency, and structural validity.

\textbf{Reinforcement Learning (RL) for LLMs.} Supervised fine-tuning adapts a VLM to the desired format but does not eliminate graph-level errors: the model may preserve valid syntax while missing relations, adding unsupported objects, or generating redundant edges~\cite{chen2025compile}. Recent work, including R1-SGG~\cite{chen2025compile}, has shown that GRPO~\cite{shao2024deepseekmath} and graph-oriented rewards improve end-to-end SGG. However, recall-oriented rewards can encourage overly complete graphs at the cost of hallucinations. In contrast, SceneGraphVLM uses a hallucination-aware reward that jointly accounts for relation coverage, precision, structural validity, and penalties for unsupported objects and edges, making training better aligned with practical deployment.

\section{Method}
\label{sec:method}

\begin{figure}[!h]
    \centering
    \includegraphics[width=1.0\linewidth]{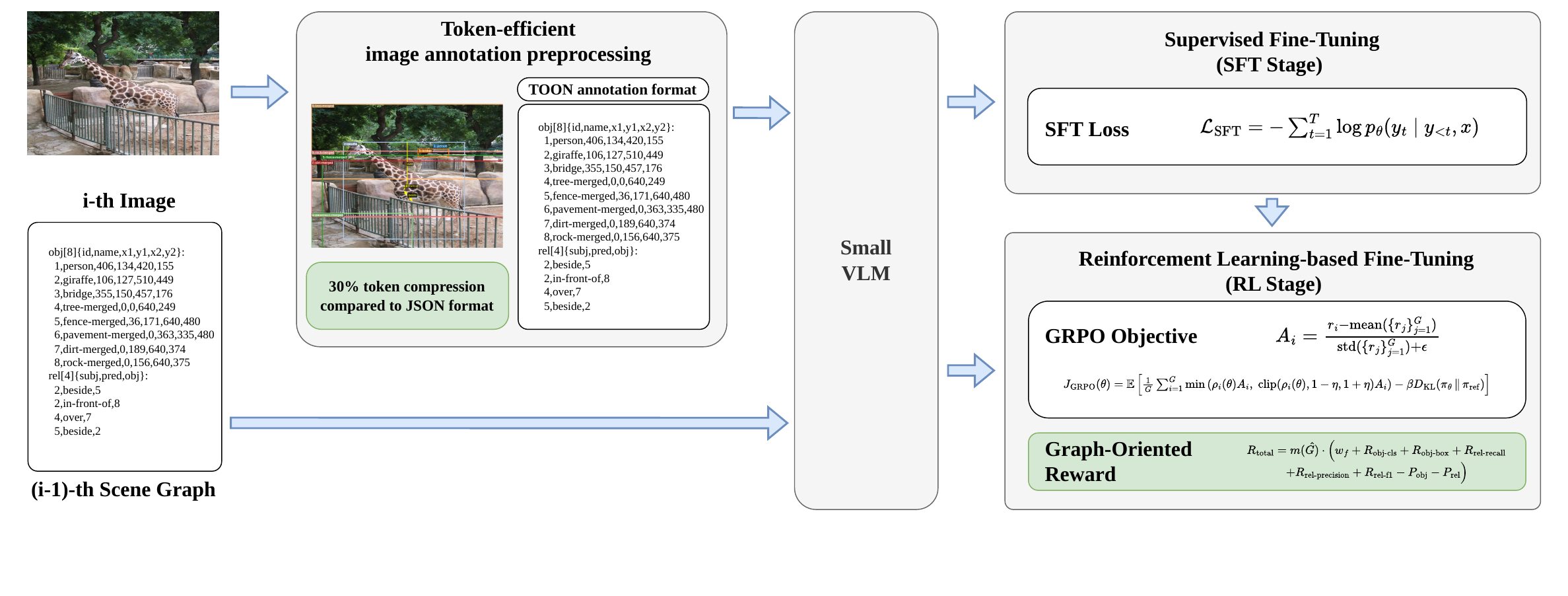}
    \caption{Overview of the SceneGraphVLM training pipeline. Image annotations are converted into a compact TOON representation and used to construct multimodal prompts. The model is first trained with supervised fine-tuning and then optimized with GRPO using hallucination-aware graph rewards over objects, bounding boxes, relations, and structural validity.}
    \label{fig:learning}
\end{figure}

The overall training pipeline of SceneGraphVLM is shown in Fig.~\ref{fig:learning}. Given an image or video frame $I_t$, the model generates a directed scene graph $G=(V,E)$, where each node $v_i \in V$ denotes an object with category $c_i$ and bounding box $b_i$, and each edge $e_{ij}=\langle v_i,p_{ij},v_j\rangle$ denotes a visual relation $p_{ij}$ between a subject and an object. For videos, the input may additionally include the previous predicted graph $\hat{G}_{t-1}$. Instead of decomposing SGG into separate detection and relation-classification modules, SceneGraphVLM treats the graph as a structured textual output and generates it directly in the TOON format. Training then proceeds in two stages: SFT aligns the VLM with the target graph schema, while GRPO optimizes graph-level rewards that account for format validity, object grounding, relation coverage, precision, and hallucination penalties.

\subsection{Token-Efficient Image Annotation Preprocessing}
\label{sec:processing}

\begin{figure}[!h]
    \centering
    \includegraphics[width=1.0\linewidth]{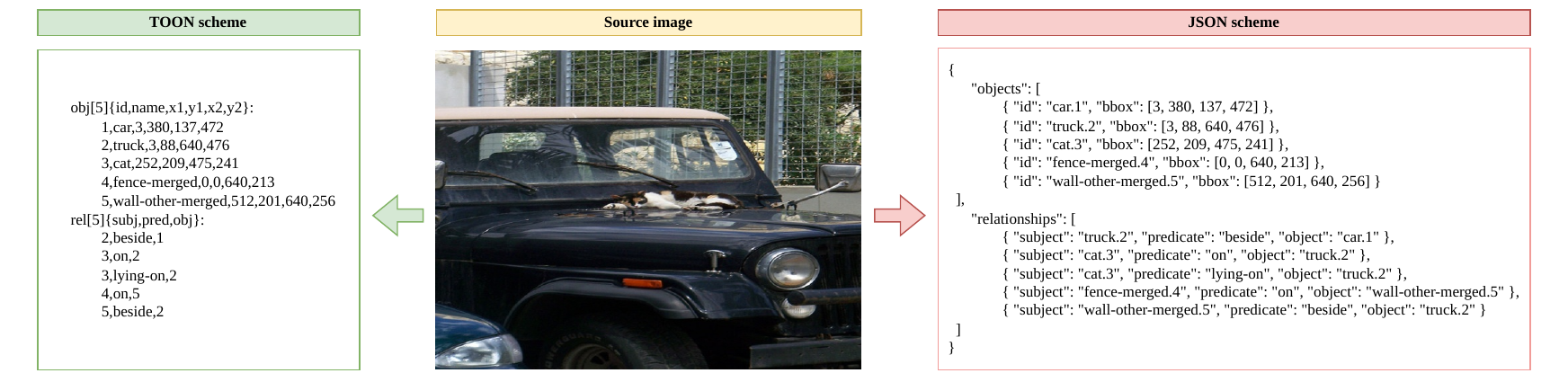}
    \caption{Same PSG scene graph represented in the compact TOON format and in JSON. TOON removes repeated keys and nested boilerplate while preserving objects, bounding boxes, and relations.}
    \label{fig:json_vs_toon}
    \vspace{-0.8em}
\end{figure}

\textbf{Compact scene graph serialization.}
Decoding JSON graphs is expensive because repeated keys, quotes, curly braces, and nested structures increase the sequence length. We therefore adapt JSON annotations to the TOON schema~\cite{Schopplich2025TOON}: objects are stored as rows under a shared header, and relations use compact dataset-specific blocks. Fig.~\ref{fig:json_vs_toon} shows the same PSG graph in both TOON and JSON formats.

\textbf{Token reduction.}
We quantify the effect of TOON in Sec.~\ref{sec:setup}.
Across PSG, PVSG, and Action Genome, TOON reduces the mean annotation length
by $1.24\times$, $1.17\times$, and $1.17\times$, respectively, while preserving
the same object, bounding-box, and relation information.

\subsection{Small Visual-Language Model}
\label{sec:small_vlm}

Since generative SGG is strongly affected by autoregressive decoding cost, we performed a preliminary backend benchmark over compact open-source VLMs. We evaluated four inference backends: Hugging Face (HF)~\cite{wolf2020huggingfacestransformersstateoftheartnatural}, vLLM~\cite{kwon2023vllm}, LMDeploy~\cite{zhang2025efficient}, and SGLang~\cite{zheng2024sglang}. All models were evaluated with the same SGG prompt, image resolution $640 \times 480$, batch size $1$, and a single NVIDIA A100 80GB GPU. Table~\ref{tab:vlm_speed_backends} reports the average decoding throughput in tokens per second.

Based on this benchmark, we use Qwen3.5-0.8B with vLLM as the primary inference and RL-sampling backend. Qwen3.5-0.8B achieves the highest vLLM throughput among the evaluated compact models, while remaining small enough for efficient fine-tuning and rollout generation. This choice is important for our setting because scene graph generation requires autoregressively decoding structured object and relation lists, and throughput directly affects both inference latency and RL training cost.

\begin{table}[!h]
\centering
\scriptsize
\renewcommand{\arraystretch}{0.9}
\setlength{\tabcolsep}{3pt}
\caption{Scene graph generation speed in tokens/s for open VLMs under different inference backends on a single A100 80GB GPU, batch size 1. Best results per column are shown in bold; best results within each model-size group are underlined.}
\label{tab:vlm_speed_backends}
\begin{tabular}{@{}lccccc@{}}
\toprule
Model & Year & vLLM & HF & LMDeploy & SGLang \\
\midrule
Qwen3.5-0.8B~\cite{qwenteam2026qwen35omnitechnicalreport}
& 2026 & \textbf{371} & 27 & -- & \underline{274} \\
SmolVLM2-500M~\cite{marafioti2025smolvlmredefiningsmallefficient}
& 2025 & 144 & \underline{28} & -- & 240 \\
SmolVLM-500M~\cite{marafioti2025smolvlmredefiningsmallefficient}
& 2024 & 317 & 25 & -- & 240 \\

\midrule
InternVL3-1B~\cite{zhu2025internvl3}
& 2025 & \underline{341} & \underline{34} & \textbf{421} & \underline{518} \\
InternVL3.5-1B~\cite{wang2025internvl35}
& 2025 & 308 & 27 & \underline{360} & 401 \\
InternVL2.5-1B~\cite{chen2024expanding}
& 2024 & 339 & 27 & 312 & \textbf{524} \\

\midrule
Qwen3.5-2B~\cite{qwenteam2026qwen35omnitechnicalreport}
& 2026 & \underline{243} & 27 & -- & 193 \\
InternVL3-2B~\cite{zhu2025internvl3}
& 2025 & 196 & 32 & 222 & 251 \\
InternVL3.5-2B~\cite{wang2025internvl35}
& 2025 & 203 & 27 & 227 & 238 \\
Qwen3-VL-2B~\cite{bai2025qwen3}
& 2025 & 195 & 27 & 228 & 245 \\
Ovis2.5-2B~\cite{lu2025ovis2}
& 2025 & 160 & \underline{33} & -- & -- \\
InternVL2.5-2B~\cite{bai2025qwen25vltechnicalreport}
& 2024 & 206 & 27 & 239 & 252 \\
Qwen2-VL-2B~\cite{wang2024qwen2}
& 2024 & 122 & 27 & \underline{251} & \underline{256} \\

\midrule
Qwen2.5-VL-3B~\cite{bai2025qwen25vltechnicalreport}
& 2025 & 128 & 20 & 145 & 160 \\
DeepSeek-VL2-tiny-3B~\cite{wu2024deepseek}
& 2024 & \underline{245} & \textbf{36} & -- & -- \\

\midrule
Qwen3.5-4B~\cite{qwenteam2026qwen35omnitechnicalreport}
& 2026 & \underline{131} & 19 & -- & 114 \\
Qwen3-VL-4B~\cite{bai2025qwen3}
& 2025 & 117 & 20 & \underline{114} & \underline{140} \\
Qwen2.5-VL-7B~\cite{bai2025qwen25vltechnicalreport}
& 2025 & 83 & \underline{27} & 84 & 94 \\
Qwen2-VL-7B~\cite{wang2024qwen2}
& 2024 & 83 & 26 & 86 & 94 \\
\bottomrule
\end{tabular}
\end{table}

\textbf{Prompt preparation.}
Each sample is converted into a multimodal instruction prompt containing the image or frame, the expected TOON graph schema, and dataset-specific vocabularies when available. PSG and PVSG share an object--relation format, while Action Genome uses a human--object schema with grouped relation types. The model is instructed to return only the final graph inside explicit answer tags. This makes the output easier to parse and reduces format drift during generation.

\begin{figure}[!h]
    \centering
    \includegraphics[width=1.0\linewidth]{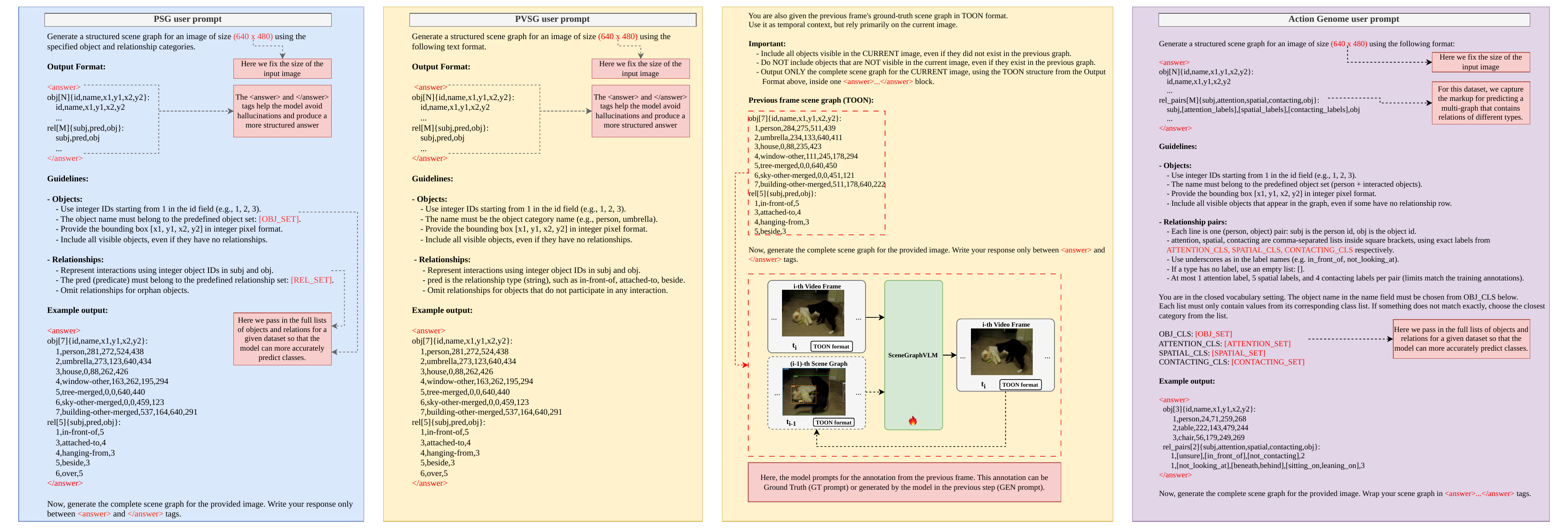}
    \caption{Dataset-specific user prompts used for SceneGraphVLM training and evaluation. PSG and PVSG use a common object--relation scene-graph format, while Action Genome uses a human--object format with attention, spatial, and contacting relation types. PVSG additionally supports temporal prompting by inserting the previous-frame scene graph, which can be either ground truth in the GT-prompt setting or model-generated in the Generated-prompt setting.}
    \label{fig:user_prompts}
\end{figure}

Figure~\ref{fig:user_prompts} summarizes the user prompts used for SFT and evaluation across PSG, PVSG, and Action Genome. All prompts ask the model to generate a complete scene graph for a resized $640 \times 480$ image. PSG uses a closed-vocabulary prompt with the full object and predicate sets. PVSG uses the same object--relation graph format, but additionally supports temporal context by inserting the previous-frame scene graph into the prompt. Action Genome requires a dataset-specific human--object format, where each human--object pair may contain attention, spatial, and contacting relations.

For video scene graph generation, we consider two temporal-context protocols.

\textbf{GT prompt.}
The model receives the ground-truth scene graph from the previous frame. This setting is not intended to represent deployment, because ground-truth previous-frame annotations are unavailable at inference time. Instead, it serves as an oracle diagnostic setting that measures the potential benefit of temporal graph context.

\textbf{Generated prompt.}
The model receives the graph generated by itself for the previous frame. This setting reflects the realistic deployment scenario. It is more challenging because errors may propagate across frames, but it better measures whether the model can use temporal graph context robustly.

In our experiments, we treat the Generated-prompt setting as the primary video evaluation protocol, while the GT-prompt setting is used as an oracle diagnostic.

\subsection{Supervised Fine-Tuning}
\label{sec:sft}

We first perform supervised fine-tuning (SFT) to align the base VLM with the scene graph
generation task and the target TOON output schema. Each training example consists
of an image or video frame, an instruction prompt, and a ground-truth scene graph
serialized in TOON. The model is trained to maximize the likelihood of the target
TOON token sequence given the multimodal input.

During this stage, optimization is performed using the standard next-token
cross-entropy loss:
\begin{equation}
\mathcal{L}_{\mathrm{SFT}}
=
-\sum_{t=1}^{T}
\log p_{\theta}\left(y_t \mid y_{<t}, x\right),
\label{eq:sft_loss}
\end{equation}
where $y_t$ denotes the ground-truth token at time step $t$, $y_{<t}$ represents
the preceding tokens in the TOON sequence, $x$ is the multimodal input consisting
of the image and instruction, and $\theta$ are the model parameters.

This stage teaches the model to follow the required schema, use the dataset
vocabulary, and produce parseable structured outputs. After SFT, the model outputs
valid TOON fairly consistently, but still produces invalid parses, missing or
spurious relations, and unstable relation coverage. We therefore further fine-tune
the SFT checkpoint with reinforcement learning, as described in the next section.

\subsection{Reinforcement Learning Fine-Tuning}
\label{sec:rl}

After SFT, the model learns to follow the required TOON schema and produces
mostly parseable scene graphs. However, token-level supervision does not directly
optimize graph-level quality metrics such as object grounding, relation coverage,
precision, and F1. We therefore further fine-tune the model with GRPO using
hallucination-aware graph rewards. The reward preserves the recall-oriented
benefits of graph-centric supervision while adding precision-oriented terms and
penalties for unsupported objects and relations. Rollout generation during RL is
accelerated with vLLM~\cite{kwon2023vllm}.

\begin{figure}[!h]
    \centering
    \includegraphics[width=1.0\linewidth]{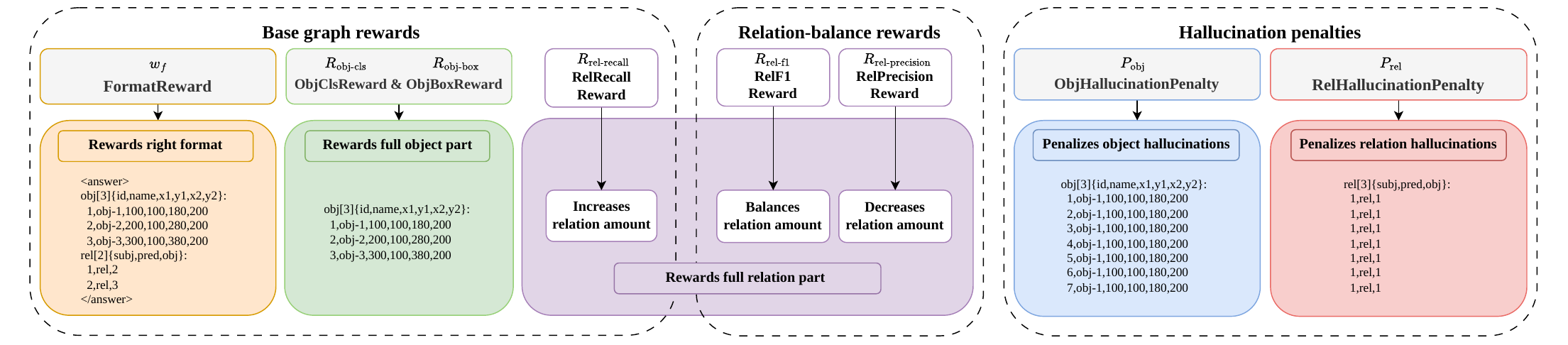}
    \caption{Overview of the hallucination-aware reward design used in SceneGraphVLM.
    The base graph rewards are adapted from the graph-centric reward design of
    R1-SGG~\cite{chen2025compile}; we extend them with relation-balance rewards
    and hallucination penalties to encourage compact and visually grounded scene graphs.}
    \label{fig:rl_rewards}
\end{figure}

Let $G=(V,E)$ and $\hat{G}=(\hat{V},\hat{E})$ denote the ground-truth and
predicted scene graphs. As shown in Fig.~\ref{fig:rl_rewards}, the total reward
is decomposed into base graph rewards, relation-balance rewards, and
hallucination penalties:
\begin{equation}
R_{\mathrm{total}}
=
m(\hat{G})
\left(
R_{\mathrm{base}}
+
R_{\mathrm{bal}}
-
P_{\mathrm{hall}}
\right),
\label{eq:total_reward_grouped}
\end{equation}
where $m(\hat{G})$ is a binary validity mask equal to one only when the generated
graph satisfies the required TOON schema. Invalid outputs receive zero reward.

\textbf{Base graph rewards.} The base graph rewards follow the general idea of R1-SGG~\cite{chen2025compile}:
they combine format validity, object-level supervision, and recall-oriented
relation matching. In our notation,
\begin{equation}
R_{\mathrm{base}}
=
w_f
+
R_{\mathrm{obj\text{-}cls}}
+
R_{\mathrm{obj\text{-}box}}
+
R_{\mathrm{rel\text{-}recall}} .
\label{eq:base_reward}
\end{equation}

The format term $w_f$ is gated by $m(\hat{G})$ and is assigned only to
structurally valid outputs. The object rewards are computed after one-to-one Hungarian matching
between predicted and ground-truth objects. For a ground-truth object
$v_i=(c_i,b_i)$ and a predicted object $\hat{v}_j=(\hat{c}_j,\hat{b}_j)$, the
matching cost is
\begin{equation}
C(v_i,\hat{v}_j)
=
\lambda_s \left(1-\operatorname{sim}(c_i,\hat{c}_j)\right)
+
\lambda_g \left(1-\operatorname{GIoU}(b_i,\hat{b}_j)\right)
+
\lambda_l \lVert b_i-\hat{b}_j\rVert_1 .
\label{eq:obj_matching_cost}
\end{equation}

Let $\pi^*$ be the resulting optimal matching.  The object classification and
localization rewards are
\begin{equation}
R_{\mathrm{obj\text{-}cls}}
=
\frac{w_{\mathrm{cls}}}{\max(|V|,|\hat{V}|)}
\sum_{(i,j)\in\pi^*}
\operatorname{sim}(c_i,\hat{c}_j),
\label{eq:obj_cls_reward}
\end{equation}
and
\begin{equation}
R_{\mathrm{obj\text{-}box}}
=
\frac{w_{\mathrm{box}}}{\max(|V|,|\hat{V}|)}
\sum_{(i,j)\in\pi^*}
\frac{
\lambda_{\mathrm{iou}}\operatorname{IoU}(b_i,\hat{b}_j)
+
\lambda_{\mathrm{l1}}\exp\left(-\lVert b_i-\hat{b}_j\rVert_1\right)
}{
\lambda_{\mathrm{iou}}+\lambda_{\mathrm{l1}}
},
\label{eq:obj_box_reward}
\end{equation}

where $sim$ is exact label equality.

For relations, we use strict one-to-one matching. A predicted relation is counted
as correct only if the subject class, predicate, and object class match the
ground truth, and both subject and object boxes satisfy the IoU threshold. Let
$M_R$ be the number of matched relations. Relation recall is
\begin{equation}
\operatorname{Recall}
=
\frac{M_R}{\max(|E|,\epsilon)},
\label{eq:rel_recall}
\end{equation}
and the recall-oriented relation reward is
\begin{equation}
R_{\mathrm{rel\text{-}recall}}
=
w_r \cdot \operatorname{Recall}.
\label{eq:rel_recall_reward}
\end{equation}

\textbf{Relation-balance rewards.} Recall-only optimization can encourage the model to increase the number of
predicted relations. To reduce this bias, we add precision- and F1-based rewards:
\begin{equation}
R_{\mathrm{bal}}
=
R_{\mathrm{rel\text{-}precision}}
+
R_{\mathrm{rel\text{-}f1}} .
\label{eq:balance_reward}
\end{equation}

Relation precision is defined as
\begin{equation}
\operatorname{Precision}
=
\frac{M_R}{\max(|\hat{E}|,\epsilon)} .
\label{eq:rel_precision}
\end{equation}

The corresponding rewards are
\begin{equation}
R_{\mathrm{rel\text{-}precision}}
=
w_p \cdot \operatorname{Precision},
\qquad
R_{\mathrm{rel\text{-}f1}}
=
w_{f1}
\cdot
\frac{
2\cdot \operatorname{Precision}\cdot \operatorname{Recall}
}{
\operatorname{Precision}+\operatorname{Recall}+\epsilon
}.
\label{eq:rel_balance_rewards}
\end{equation}

These terms encourage the model to preserve useful relation coverage without
overfilling the graph with unsupported edges.

\textbf{Hallucination penalties.} Finally, we penalize unmatched predicted objects and relations:
\begin{equation}
P_{\mathrm{hall}}
=
P_{\mathrm{obj}}
+
P_{\mathrm{rel}} .
\label{eq:hallucination_penalty}
\end{equation}

Let $M_V$ and $M_R$ be the number of matched objects and relations. The penalties
are defined as
\begin{equation}
P_{\mathrm{obj}}
=
w_{\mathrm{obj\text{-}h}}
\left(
\frac{|\hat{V}|-M_V}{\max(|\hat{V}|,\epsilon)}
\right)^{\alpha_{\mathrm{obj}}},
\qquad
P_{\mathrm{rel}}
=
w_{\mathrm{rel\text{-}h}}
\left(
\frac{|\hat{E}|-M_R}{\max(|\hat{E}|,\epsilon)}
\right)^{\alpha_{\mathrm{rel}}}.
\label{eq:hallucination_penalties}
\end{equation}

These penalties discourage ungrounded objects and spurious relations.

\section{Experimental Setup}
\label{sec:setup}

\textbf{Datasets and Preprocessing.}
We train and evaluate SceneGraphVLM on three scene graph benchmarks: PSG~\cite{yang2022psg}, PVSG~\cite{yang2023pvsg}, and Action Genome (AG)~\cite{ji2020action}. PSG provides panoptic scene graphs for static images, PVSG extends this setting to videos with frame-level annotations, and AG focuses on human--object interactions in video frames. All images and frames are resized to $640 \times 480$, and all annotations are converted into the TOON format described in Sec.~\ref{sec:processing}.

We apply a unified preprocessing pipeline that removes duplicate objects and relations, self-relations, and samples with zero ground-truth relations. The latter prevents the hallucination-aware reward from encouraging degenerate empty-relation outputs. Table~\ref{tab:dataset_cleaning} summarizes the number of examples before and after removing zero-relation samples.

\begin{table}[!h]
\centering
\scriptsize
\renewcommand{\arraystretch}{0.96}
\setlength{\tabcolsep}{4.2pt}
\caption{
Zero-relation filtering statistics. We report the number of examples before and after removing samples with no ground-truth relations, together with the removed fraction.
}
\label{tab:dataset_cleaning}
\begin{tabular}{@{}llrrrc@{}}
\toprule
\textbf{Dataset} & \textbf{Split} & \textbf{Before} & \textbf{Removed} & \textbf{Kept} & \textbf{Removed (\%)} \\
\midrule
\multirow{2}{*}{PSG}
& train & 46\,563 & 866 & 45\,697 & 1.86 \\
& test  & 2\,186  & 9   & 2\,177  & 0.41 \\
\midrule
\multirow{2}{*}{PVSG all}
& train & 126\,865 & 11\,226 & 115\,639 & 8.85 \\
& test  & 22\,604  & 1\,519  & 21\,085  & 6.72 \\
\midrule
\multirow{2}{*}{PVSG PSFR}
& train & 16\,047 & 2\,852 & 13\,195 & 17.77 \\
& test  & 3\,641  & 602   & 3\,039  & 16.53 \\
\midrule
\multirow{2}{*}{PVSG MaxInfo}
& train & 39\,056 & 3\,675 & 35\,381 & 9.41 \\
& test  & 7\,344  & 685   & 6\,659  & 9.33 \\
\midrule
\multirow{2}{*}{PVSG BaseAnnot}
& train & 30\,297 & 5\,150 & 25\,147 & 17.00 \\
& test  & 6\,405  & 811   & 5\,594  & 12.66 \\
\bottomrule
\end{tabular}
\end{table}

\textbf{TOON token statistics.}
To quantify serialization efficiency, we serialize the same annotations in JSON and TOON
and measure token lengths with the Qwen3.5-0.8B tokenizer.
Table~\ref{tab:toon_token_stats_compact} reports token statistics for all datasets.
Compression is defined as the JSON token length divided by the TOON token length,
so values larger than $1$ indicate shorter TOON outputs.
TOON consistently reduces both mean and median annotation length across PSG, PVSG,
and Action Genome. In a few very short examples, TOON can be slightly longer than JSON
because fixed headers introduce a small constant overhead.

\begin{table}[!h]
\centering
\tiny
\renewcommand{\arraystretch}{1.08}
\setlength{\tabcolsep}{5.2pt}
\caption{
Token-length statistics for JSON and TOON annotations using the Qwen3.5-0.8B tokenizer.
For TOON rows, green subscripts show token reduction relative to JSON for the same statistic.
AG denotes Action Genome.
}
\label{tab:toon_token_stats_compact}
\resizebox{0.70\textwidth}{!}{%
\begin{tabular}{@{}llcccc@{}}
\toprule
\textbf{Dataset}
& \textbf{Format}
& \textbf{Min Tokens $\downarrow$}
& \textbf{Mean Tokens $\downarrow$}
& \textbf{Median Tokens $\downarrow$}
& \textbf{Max Tokens $\downarrow$} \\
\midrule

\multirow{2}{*}{PSG}
& JSON & 69 & 399 & 332 & 2273 \\
& TOON
& $64_{\scriptscriptstyle \textcolor{green!50!black}{(-7\%)}}$
& $323_{\scriptscriptstyle \textcolor{green!50!black}{(-19\%)}}$
& $260_{\scriptscriptstyle \textcolor{green!50!black}{(-22\%)}}$
& $1947_{\scriptscriptstyle \textcolor{green!50!black}{(-14\%)}}$ \\

\midrule
\multirow{2}{*}{PVSG}
& JSON & 67 & 320 & 303 & 1203 \\
& TOON
& $62_{\scriptscriptstyle \textcolor{green!50!black}{(-7\%)}}$
& $273_{\scriptscriptstyle \textcolor{green!50!black}{(-15\%)}}$
& $257_{\scriptscriptstyle \textcolor{green!50!black}{(-15\%)}}$
& $954_{\scriptscriptstyle \textcolor{green!50!black}{(-21\%)}}$ \\

\midrule
\multirow{2}{*}{AG}
& JSON & 49 & 153 & 154 & 621 \\
& TOON
& $26_{\scriptscriptstyle \textcolor{green!50!black}{(-47\%)}}$
& $131_{\scriptscriptstyle \textcolor{green!50!black}{(-14\%)}}$
& $131_{\scriptscriptstyle \textcolor{green!50!black}{(-15\%)}}$
& $430_{\scriptscriptstyle \textcolor{green!50!black}{(-31\%)}}$ \\

\bottomrule
\end{tabular}%
}
\end{table}

For video experiments, we use previous-frame scene graphs as optional temporal context. During SFT, this context can be the ground-truth graph from the previous frame, which helps the model learn local temporal continuity. However, this creates a train--inference mismatch: in deployment, the previous-frame graph is generated by the model itself, so early mistakes can propagate through the sequence. To reduce this dependence, RL fine-tuning uses corrupted previous-frame graphs in the prompt, making the policy less sensitive to imperfect temporal context and more robust under generated-graph inference.

\textbf{Video Frame Thinning.}
PVSG contains many adjacent frames with nearly identical content and scene graphs, making full-frame training redundant and prone to overfitting. We therefore use frame thinning for both training and evaluation. For training, we use the annotation-based BaseAnnot split. For evaluation, we report results on the PSFR-selected test split~\cite{zemskova2026focusgraph}. As shown in Table~\ref{tab:pvsg_thinning}, PSFR keeps $12.6\%$ of training frames and $16.1\%$ of test frames while preserving full or near-full object-category and predicate vocabulary coverage.

We evaluate three frame-selection strategies:
\textit{BaseAnnot}, which keeps frames when object categories or relation counts change in the annotation;
\textit{MaxInfo}, a training-free embedding-space selection method;
and \textit{PSFR}, a training-free key-frame selector based on patchwise sparse-flow retention.
Table~\ref{tab:pvsg_thinning} reports the number of retained frames and the corresponding object and predicate vocabulary coverage before zero-relation filtering.

\begin{table}[!h]
\centering
\scriptsize
\renewcommand{\arraystretch}{0.98}
\setlength{\tabcolsep}{4.2pt}
\caption{
PVSG frame selection before zero-relation filtering.
Object and predicate columns report category coverage before $\to$ after selection.
PSFR provides the strongest frame reduction, while MaxInfo preserves the most predicate categories.
}
\label{tab:pvsg_thinning}
\begin{tabular}{@{}llrrrcc@{}}
\toprule
\textbf{Split}
& \textbf{Method}
& \textbf{Frames before}
& \textbf{Frames after}
& \textbf{\% kept}
& \textbf{Objects}
& \textbf{Predicates} \\
\midrule
\multirow{3}{*}{train}
& PSFR
& 126\,865 & 16\,047 & \textbf{12.6}
& \textbf{125$\to$125} & 61$\to$60 \\
& MaxInfo
& 126\,865 & 39\,056 & 30.8
& \textbf{125$\to$125} & \textbf{61$\to$61} \\
& BaseAnnot
& 126\,865 & 30\,297 & 23.9
& 125$\to$107 & 61$\to$54 \\
\midrule
\multirow{3}{*}{test}
& PSFR
& 22\,604 & 3\,641 & \textbf{16.1}
& 108$\to$106 & 58$\to$56 \\
& MaxInfo
& 22\,604 & 7\,344 & 32.5
& 108$\to$106 & \textbf{58$\to$58} \\
& BaseAnnot
& 22\,604 & 6\,405 & 28.3
& \textbf{108$\to$107} & 58$\to$54 \\
\bottomrule
\end{tabular}
\end{table}

\textbf{Baselines.}
We compare SceneGraphVLM against three groups of baselines. First, we include R1-SGG~\cite{chen2025compile}, an end-to-end structured-output SGG method trained with GRPO-style graph rewards. Second, we evaluate open-source VLMs in the same prompt-based SGG setting. Third, we evaluate commercial multimodal models, including GPT-5.4 and Gemini-3 family models, in a zero-shot setting. For fair comparison, model outputs are parsed into a common graph representation and evaluated with the same protocol. On AG, we additionally compare under the standard SGDET-style evaluation, extended with precision-oriented metrics. All open-source models and SGG pipelines were evaluated on a single Nvidia A100 80GB GPU.

\textbf{Metrics.}
Classical SGG evaluation commonly relies on Recall@K over ranked relation
triplets. This protocol is well suited to detector-based models with confidence
scores, but is less natural for generative VLMs that output complete scene graphs
as structured text. It can also favor over-generation, since producing many
unsupported objects and relations may improve recall. We therefore report
object- and relation-level Precision, Recall, and F1, and define the final SGG
score as the mean of object F1 and relation F1. Metrics are computed per sample
and then macro-averaged; therefore the reported F1 is not necessarily the
harmonic mean of the aggregate Precision and Recall.

For PSG and PVSG, strict lexical matching is used as the primary evaluation
setting. We additionally report an LLM-assisted soft matching protocol for
semantically equivalent labels, such as \texttt{person} versus \texttt{man} or
\texttt{on} versus \texttt{parked-on}. Objects are first aligned by bounding-box
IoU using Hungarian matching~\cite{kuhn1955hungarian}; exact matches are counted
directly, while disputed aligned object names and predicates are resolved by a
Qwen judge. This protocol is conceptually related to LLM-based evaluation methods
such as GPTScore~\cite{fu2024gptscore}, but the LLM is used only to resolve
ambiguous semantic matches, while final metrics are still computed from TP, FP,
and FN counts.

\begin{figure}[!h]
    \centering
    \includegraphics[width=0.9\linewidth]{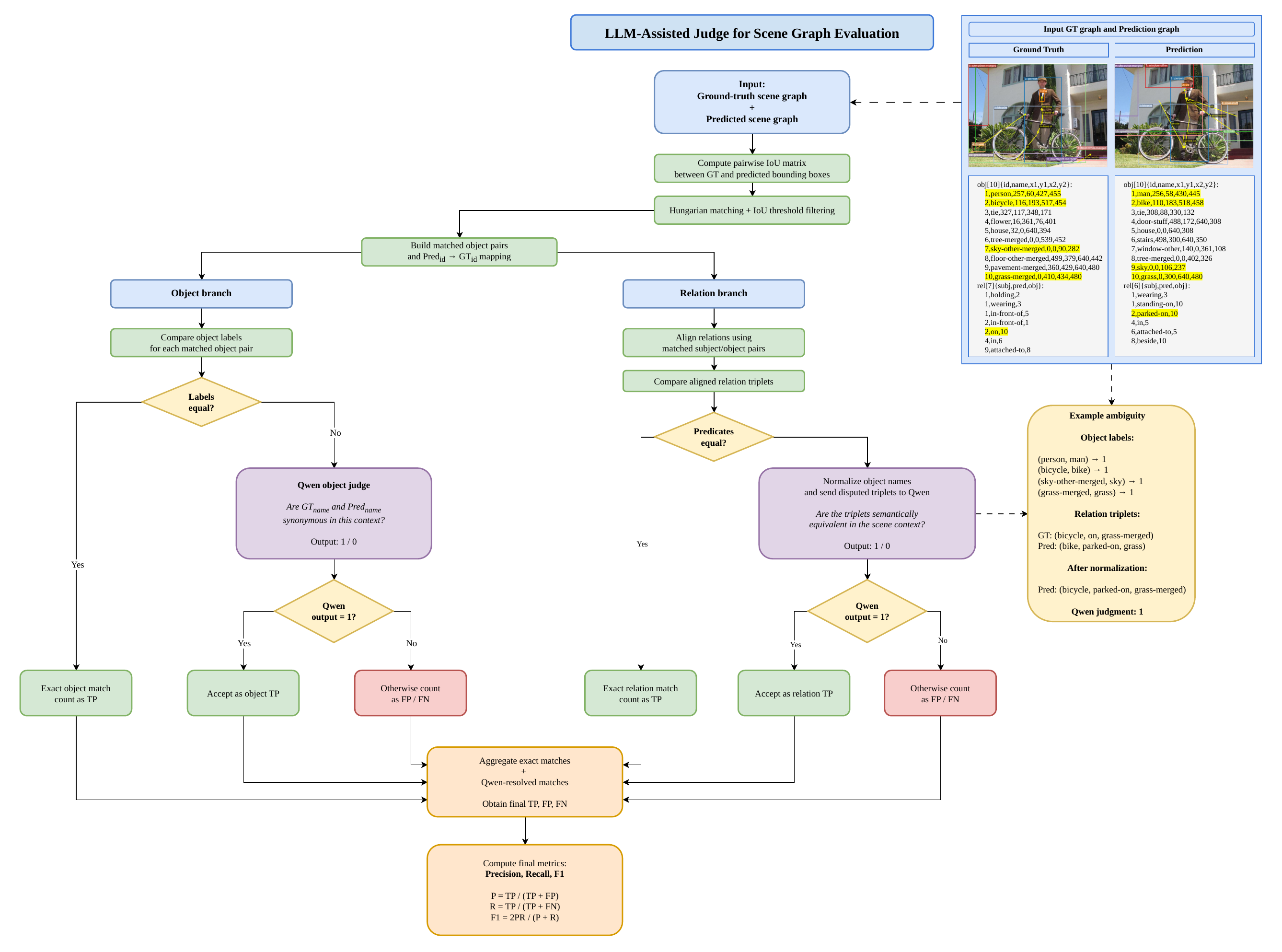}
    \caption{LLM-assisted evaluation protocol. Objects are first matched by bounding-box IoU using Hungarian matching. Exact object labels and relation predicates are counted directly, while disputed object names and predicates are resolved by a Qwen judge, instantiated as \texttt{Qwen3-4B-Instruct-2507}, with access to the full scene context. Exact and judge-resolved matches are aggregated into TP, FP, and FN for Precision, Recall, and F1.}
    \label{fig:qwen_metrics}
\end{figure}

Figure~\ref{fig:qwen_metrics} illustrates the LLM-assisted evaluation protocol used for PSG and PVSG. The protocol is designed for generative VLMs that output complete scene graphs rather than ranked triplet lists. First, predicted and ground-truth objects are aligned by bounding-box IoU using Hungarian matching. Exact object-label matches are accepted directly. When the labels differ but the boxes are aligned, the object pair is treated as disputed and passed to an LLM judge together with the full scene context.

In our implementation, the judge is instantiated as \texttt{Qwen3-4B-Instruct-2507}. The judge is not used as a free-form graph evaluator: it is queried only after deterministic geometric matching and exact lexical matching have been applied. For object labels, it decides whether two aligned names are semantically equivalent in context. For relations, it is queried only for non-identical predicates whose subject and object have already been aligned through the object-matching step.

Relation evaluation is then performed on top of the aligned object pairs. Exact predicate matches are counted directly, while disputed predicates are judged for contextual semantic equivalence. The final set of exact and judge-resolved matches is aggregated into TP, FP, and FN for object-level and relation-level Precision, Recall, and F1.

For Action Genome, we follow the standard SGDET with-constraint protocol and
additionally report Precision@K and F1@K over the same matched triplets. These
precision-oriented metrics make the evaluation sensitive to hallucinated
relations and better reflect compact generative scene graph outputs.

\textbf{Training Details.}
All SceneGraphVLM models use Qwen3.5-0.8B~\cite{qwenteam2026qwen35omnitechnicalreport} as the backbone and are trained with SWIFT~\cite{zhao2025swift} using full fine-tuning unless stated otherwise.

\paragraph{SFT stage.}
The SFT stage is optimized with AdamW through SWIFT with the following settings:
bfloat16 precision, FlashAttention~\cite{dao2023flashattention}, DeepSpeed ZeRO-2~\cite{rajbhandari2020zero}, maximum sequence length
$8192$, learning rate $10^{-5}$, warmup ratio $0.05$, per-device batch size $4$,
gradient accumulation steps $2$, and group-by-length batching.
We train for $2$ epochs and select the best checkpoint by validation loss.
All runs are logged with Comet ML~\cite{cometml_tracking_2022}.

\paragraph{RL stage.}
The best SFT checkpoint is used as initialization for GRPO fine-tuning with the
hallucination-aware reward described in Sec.~\ref{sec:rl}.
We use the SWIFT RLHF interface with vLLM in colocated mode for accelerated
rollout generation. Only the LLM backbone is updated; the visual encoder and
aligner are frozen (\texttt{freeze\_vit=true}, \texttt{freeze\_aligner=true}).

Key GRPO hyperparameters are as follows: learning rate $6\times10^{-7}$, KL coefficient
$\beta=0.04$, clip range $[\epsilon, \epsilon_{\mathrm{high}}]=[0.20,\,0.28]$,
$G=12$ rollouts per sample, per-device train batch size $24$, generation batch
size $48$, gradient accumulation $1$, and no reward scaling.
Sequences are truncated to $2048$ input and $1024$ completion tokens;
samples exceeding the limit are discarded (\texttt{truncation\_strategy=delete}).
vLLM is configured with GPU memory utilization $0.30$, maximum model length
$4096$, and tensor parallel size $1$. All other settings follow the SFT stage
(bfloat16, FlashAttention-2~\cite{dao2023flashattention}, DeepSpeed ZeRO-2~\cite{rajbhandari2020zero}, max grad norm $1.0$).

The active reward functions and their weights are listed in
Table~\ref{tab:reward_weights}. Five diagnostic signals
(\textit{frac\_no\_rel}, \textit{num\_pred\_objs}, \textit{num\_pred\_rels},
\textit{frac\_invalid\_rel}, \textit{has\_answer\_tags}) are logged for
monitoring but assigned zero weight and do not affect training.
The best checkpoint is selected by mean reward evaluated every $100$ steps.

\begin{table}[h]
\centering
\scriptsize
\setlength{\tabcolsep}{5pt}
\caption{Reward functions used during GRPO fine-tuning.
The GRPO combination weight is $1$ for all active terms (set via \texttt{--reward\_weights}).
Internal weights are baked into each ORM and scale the raw score before GRPO aggregation.}
\label{tab:reward_weights}
\begin{tabular}{@{}llcc@{}}
\toprule
\textbf{Group} & \textbf{Reward function} & \textbf{Weight} & \textbf{Sign} \\
\midrule
Format
  & \textit{FormatReward}             & 0.5  & $+$ \\
\midrule
\multirow{2}{*}{Object}
  & \textit{ObjclsReward}            & 1.5  & $+$ \\
  & \textit{ObjBoxReward}            & 1.5  & $+$ \\
\midrule
\multirow{3}{*}{Relation}
  & \textit{RelRecallReward}           & 3.0  & $+$ \\
  & \textit{RelPrecisionReward}      & 1.0  & $+$ \\
  & \textit{RelF1Reward}             & 2.0  & $+$ \\
\midrule
\multirow{2}{*}{Hallucination}
  & \textit{ObjHallucinationPenalty}  & 1.0\textsuperscript{$\dagger$}  & $-$ \\
  & \textit{RelHallucinationPenalty} & 1.0\textsuperscript{$\dagger$} & $-$ \\
\bottomrule
\multicolumn{4}{@{}l@{}}{\textsuperscript{$\dagger$}Applied as
  $w \cdot \bigl(\text{FP fraction}\bigr)^{2}$; exponent $\alpha=2$.} \\
\end{tabular}
\end{table}

\paragraph{Hardware and training time.}
All RL experiments are run on two NVIDIA H200 GPUs.
One epoch of GRPO fine-tuning takes approximately $1$ day on PVSG (PSFR split), $2$ days on PSG, and $4$ days on Action Genome. Dataset-specific prompt templates and video temporal-context protocols are described in Sec.~\ref{sec:small_vlm} and shown in Fig.~\ref{fig:user_prompts}. Inference and RL sampling use vLLM for accelerated generation.

\section{Experimental Results}
\label{sec:results}

\subsection{Comparison with Existing Methods}

\textbf{PSG results.}
Table~\ref{tab:psg_toon_iou50_combined} reports the full results on the static-image PSG benchmark at IoU threshold $50$.
Zero-shot commercial VLMs produce structurally valid graphs more reliably than open-source base models in our prompting setup, but remain slower and substantially less accurate than the fine-tuned SceneGraphVLM.
Gemini-3-flash reaches the strongest commercial result with a Soft SGG score of $0.318$, but requires $4.67$\,s per graph.
Base Qwen models are faster, but their relation quality remains weak, especially for Qwen3.5-2B, whose relation F1-score is close to zero.

After supervised fine-tuning, SceneGraphVLM improves both object and relation quality by a large margin.
On PSG, it achieves the best SGG score among reported models, $0.442$ under strict matching and $0.460$ under LLM-assisted semantic evaluation, while keeping generation time below one second and producing no parsing failures.
On PSG, SceneGraphVLM operates in the sub-second regime, generating complete scene graphs in $0.643$\,s on average while outperforming larger commercial and generative SGG baselines.
This suggests that compact TOON outputs and task-specific fine-tuning are essential for reliable VLM-based scene graph generation.

\begin{figure}[!h]
    \centering
    \includegraphics[width=1.0\linewidth]{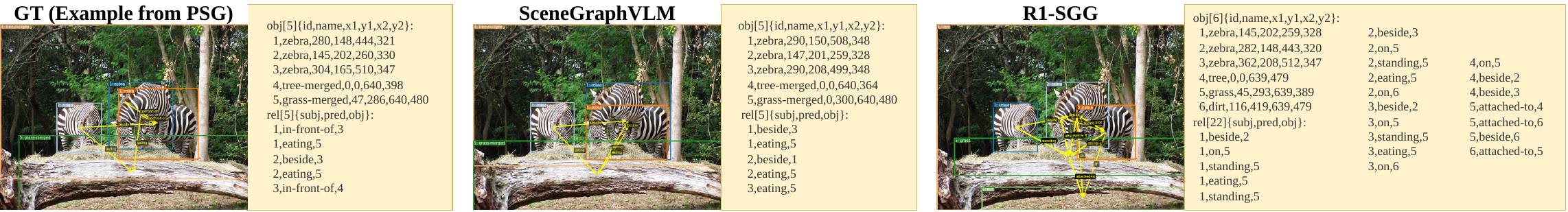}
    \caption{
    Qualitative PSG comparison between the ground-truth graph, R1-SGG, and SceneGraphVLM.
    SceneGraphVLM produces a more compact and grounded graph, while R1-SGG generates a denser recall-oriented prediction.
    }
    \label{fig:psg_qualitative_comparison}
\end{figure}

Figure~\ref{fig:psg_qualitative_comparison} provides a qualitative comparison between the ground-truth graph, R1-SGG, and SceneGraphVLM on PSG.
R1-SGG recovers several relevant objects, but its prediction is strongly recall-oriented: it generates a denser graph with redundant or weakly grounded relations, including repeated spatial and support predicates between similar objects and background regions.
In contrast, SceneGraphVLM produces a compact graph whose number of objects and relations is closer to the ground truth.
It preserves the main visually supported relations, such as interactions between zebras and grass, while avoiding relation over-generation.
This behavior illustrates the effect of hallucination-aware training: the model favors precise, grounded scene graphs over adding relations only to increase coverage.

\begin{table}[!h]
\centering
\scriptsize
\renewcommand{\arraystretch}{1.12}
\setlength{\tabcolsep}{2.2pt}
\caption{
Full results on the PSG dataset in TOON format at IoU threshold $=50$.
We report generation time, failure rate, object metrics, relation metrics, and the final SGG score
under both Strict and Soft evaluation.
Best overall results are shown in bold; best results within each model group are underlined.
}
\label{tab:psg_toon_iou50_combined}

\resizebox{\textwidth}{!}{%
\begin{tabular}{@{}l c c c cc cc cc cc cc cc cc@{}}
\toprule
\multirow{2}{*}{\textbf{Model}}
& \multirow{2}{*}{\textbf{Mode}}
& \multirow{2}{*}{\textbf{\makecell{Fail.\\rate (\%)}}}
& \multirow{2}{*}{\textbf{\makecell{Gen. time\\ (s)}}}
& \multicolumn{2}{c}{\textbf{Obj Prec}}
& \multicolumn{2}{c}{\textbf{Obj Rec}}
& \multicolumn{2}{c}{\textbf{Obj F1}}
& \multicolumn{2}{c}{\textbf{Rel Prec}}
& \multicolumn{2}{c}{\textbf{Rel Rec}}
& \multicolumn{2}{c}{\textbf{Rel F1}}
& \multicolumn{2}{c}{\textbf{SGG Score}} \\
\cmidrule(lr){5-6}
\cmidrule(lr){7-8}
\cmidrule(lr){9-10}
\cmidrule(lr){11-12}
\cmidrule(lr){13-14}
\cmidrule(lr){15-16}
\cmidrule(l){17-18}
& & &
& \textbf{Strict} & \textbf{Soft}
& \textbf{Strict} & \textbf{Soft}
& \textbf{Strict} & \textbf{Soft}
& \textbf{Strict} & \textbf{Soft}
& \textbf{Strict} & \textbf{Soft}
& \textbf{Strict} & \textbf{Soft}
& \textbf{Strict} & \textbf{Soft} \\
\midrule

\multicolumn{18}{@{}l@{}}{\textit{\textbf{Commercial M-LLMs}}} \\

GPT-5.4-nano
& Base & 0.64 & 2.456$\pm$0.527
& 0.075 & 0.148 & 0.059 & 0.124
& 0.062 & 0.126
& 0.013 & 0.017 & 0.007 & 0.009 & 0.008 & 0.011
& 0.035 & 0.069 \\

GPT-5-nano
& Base & 0.78 & 32.639$\pm$5.672
& 0.078 & 0.138 & 0.063 & 0.116
& 0.065 & 0.117
& 0.011 & 0.015 & 0.007 & 0.009 & 0.008 & 0.011
& 0.036 & 0.064 \\

GPT-5-mini
& Base & 0.32 & 26.650$\pm$3.977
& 0.075 & 0.155 & 0.097 & 0.203
& 0.080 & 0.166
& 0.020 & 0.026 & 0.037 & 0.049 & 0.024 & 0.032
& 0.052 & 0.099 \\

GPT-5.4-mini
& Base & 0.74 & \underline{2.213$\pm$0.546}
& 0.234 & 0.325 & 0.167 & 0.244
& 0.186 & 0.265
& 0.058 & 0.074 & 0.032 & 0.042 & 0.038 & 0.050
& 0.112 & 0.157 \\

Gemini-3-flash
& Base & \underline{0.09} & 4.671$\pm$0.723
& \underline{0.312} & \underline{0.497}
& \underline{0.303} & \underline{0.509}
& \underline{0.292} & \underline{0.477}
& \underline{0.116} & \underline{0.156}
& \underline{0.128} & \underline{0.174}
& \underline{0.117} & \underline{0.158}
& \underline{0.204} & \underline{0.318} \\

\midrule
\multicolumn{18}{@{}l@{}}{\textit{\textbf{Open-source M-LLMs}}} \\

SmolVLM-500M-Instruct
& Base & 86.08 & 2.871$\pm$0.711
& 0.003 & 0.003 & 0.001 & 0.001
& 0.001 & 0.001
& 0.000 & 0.000 & 0.000 & 0.000 & 0.000 & 0.000
& 0.001 & 0.001 \\

InternVL2.5-1B
& Base & 6.85 & 1.169$\pm$0.068
& 0.033 & 0.044 & 0.010 & 0.014
& 0.014 & 0.020
& 0.000 & 0.000 & 0.000 & 0.000 & 0.000 & 0.000
& 0.007 & 0.010 \\

DeepSeek-VL2-tiny-3B
& Base & \underline{0.64} & 3.747$\pm$0.622
& 0.058 & 0.071 & 0.017 & 0.021
& 0.025 & 0.030
& 0.000 & 0.000 & 0.000 & 0.000 & 0.000 & 0.000
& 0.012 & 0.015 \\

InternVL3-1B
& Base & 11.40 & 2.669$\pm$0.756
& 0.105 & 0.124 & 0.034 & 0.041
& 0.047 & 0.057
& 0.003 & 0.005 & 0.001 & 0.001 & 0.001 & 0.002
& 0.024 & 0.030 \\

Ovis2.5-2B-Instruct
& Base & 75.97 & 5.738$\pm$0.413
& 0.070 & 0.104 & 0.049 & 0.078
& 0.054 & 0.084
& 0.013 & 0.018 & 0.004 & 0.007 & 0.006 & 0.009
& 0.030 & 0.047 \\

Qwen3.5-0.8B
& Base & 67.92 & \underline{1.074$\pm$0.412}
& 0.140 & 0.164 & 0.045 & 0.055
& 0.063 & 0.075
& 0.000 & 0.001 & 0.000 & 0.000 & 0.000 & 0.000
& 0.032 & 0.038 \\

InternVL3.5-1B
& Base & 23.81 & 2.981$\pm$0.556
& 0.105 & 0.124 & 0.034 & 0.041
& 0.084 & 0.153
& 0.007 & 0.014 & 0.004 & 0.008 & 0.005 & 0.010
& 0.045 & 0.081 \\

Qwen3-VL-4B
& Base & 45.37 & 7.778$\pm$0.613
& 0.362 & 0.379 & 0.078 & 0.086
& 0.120 & 0.129
& 0.007 & 0.009 & 0.002 & 0.003 & 0.003 & 0.004
& 0.061 & 0.067 \\

Qwen3-VL-2B
& Base & 50.09 & 7.460$\pm$0.563
& 0.257 & 0.277 & 0.110 & 0.125
& 0.142 & 0.158
& 0.020 & 0.033 & 0.005 & 0.009 & 0.008 & 0.013
& 0.075 & 0.085 \\

Qwen3.5-2B
& Base & 14.61 & 1.640$\pm$0.441
& 0.230 & 0.319 & 0.187 & \underline{0.272}
& 0.194 & 0.276
& 0.004 & 0.005 & 0.001 & 0.002 & 0.002 & 0.003
& 0.098 & 0.139 \\

Qwen3.5-4B
& Base & 7.58 & 3.042$\pm$0.421
& \underline{0.399} & \underline{0.474}
& \underline{0.217} & 0.271
& \underline{0.265} & \underline{0.323}
& \underline{0.080} & \underline{0.103}
& \underline{0.037} & \underline{0.047}
& \underline{0.045} & \underline{0.058}
& \underline{0.155} & \underline{0.190} \\

\midrule
\multicolumn{18}{@{}l@{}}{\textit{\textbf{Fine-tuned non-final variants}}} \\

\makecell[l]{InternVL3.5-1B-Instruct\\(LoRA $r{=}16$, $\alpha{=}32$)}
& \makecell{SFT\\close}
& 10.34 & 1.003$\pm$0.286
& 0.249 & 0.255 & 0.218 & 0.223
& 0.227 & 0.233
& 0.080 & 0.093 & 0.040 & 0.047 & 0.050 & 0.059
& 0.139 & 0.146 \\

\makecell[l]{InternVL3-1B\\(full finetune LLM head)}
& \makecell{SFT\\open}
& 1.24 & 0.916$\pm$0.253
& 0.339 & 0.348 & 0.313 & 0.321
& 0.321 & 0.329
& 0.131 & 0.145 & 0.071 & 0.080 & 0.087 & 0.098
& 0.204 & 0.213 \\

\makecell[l]{InternVL3-1B\\(full finetune LLM head)}
& \makecell{SFT\\close}
& 0.78 & \underline{0.894$\pm$0.236}
& \underline{0.360} & \underline{0.368}
& 0.321 & 0.328
& 0.334 & 0.342
& \underline{0.135} & 0.150
& 0.080 & 0.090
& 0.095 & 0.107
& 0.215 & 0.224 \\

\makecell[l]{InternVL3.5-1B-Instruct\\(full finetune LLM head)}
& \makecell{SFT\\close}
& \underline{0.37} & 1.115$\pm$0.346
& 0.341 & 0.350
& \underline{0.337} & \underline{0.346}
& \underline{0.335} & \underline{0.343}
& 0.134 & \underline{0.151}
& \underline{0.085} & \underline{0.097}
& \underline{0.098} & \underline{0.111}
& \underline{0.216} & \underline{0.227} \\

\midrule
\multicolumn{18}{@{}l@{}}{\textit{\textbf{InternVL2.5-1B variants}}} \\

\makecell[l]{SceneGraphVLM\\(InternVL2.5-1B)}
& \makecell{SFT\\close}
& 3.45 & \underline{0.704$\pm$0.210}
& \underline{0.629} & \underline{0.641}
& 0.438 & 0.446
& \underline{0.506} & 0.516
& \underline{0.230} & \underline{0.277}
& 0.111 & 0.137
& 0.142 & 0.174
& 0.324 & 0.345 \\

\makecell[l]{SceneGraphVLM\\(InternVL2.5-1B)}
& \makecell{SFT+RL\\close}
& \underline{1.15} & \underline{0.704$\pm$0.210}
& 0.610 & 0.624
& \underline{0.446} & \underline{0.456}
& \underline{0.506} & \underline{0.517}
& 0.222 & 0.275
& \underline{0.114} & \underline{0.145}
& \underline{0.143} & \underline{0.182}
& \underline{0.325} & \underline{0.349} \\

\midrule

\makecell[l]{R1-SGG~\cite{chen2025compile}}
& \makecell{SFT+RL\\close}
& 0.14 & 8.564$\pm$0.580
& 0.476 & 0.610
& 0.380 & 0.574
& 0.410 & 0.592
& 0.105 & 0.117
& \textbf{0.364} & \textbf{0.405}
& 0.158 & 0.176
& 0.284 & 0.386 \\

\makecell[l]{SceneGraphVLM}
& \makecell{SFT\\close}
& \textbf{0.00} & \textbf{0.643$\pm$0.191}
& 0.632 & 0.642
& 0.589 & 0.598
& 0.602 & 0.611
& 0.272 & 0.304
& 0.194 & 0.219
& 0.213 & 0.239
& 0.408 & 0.425 \\

\makecell[l]{SceneGraphVLM}
& \makecell{SFT + RL\\close}
& \textbf{0.00} & \textbf{0.643$\pm$0.191}
& \textbf{0.679} & \textbf{0.690}
& \textbf{0.601} & \textbf{0.611}
& \textbf{0.630} & \textbf{0.640}
& \textbf{0.338} & \textbf{0.372}
& 0.217 & 0.241
& \textbf{0.253} & \textbf{0.280}
& \textbf{0.442} & \textbf{0.460} \\

\bottomrule
\end{tabular}%
}
\end{table}

\textbf{PVSG results.}
Table~\ref{tab:pvsg_iou50_label_modes_combined} summarizes the full results on PVSG under two temporal-prompt settings.
In the GT-prompt setting, the previous-frame graph is taken from ground truth and therefore provides clean temporal context.
Commercial VLMs perform reasonably well, with Gemini-3-flash reaching a Strict-based SGG score of $0.419$, but they remain slower than SceneGraphVLM.
R1-SGG and TRaSER perform substantially worse under our evaluation protocol, especially on relation metrics.

SceneGraphVLM achieves the strongest PVSG performance in both prompt settings.
With GT previous-frame context, RL improves the Strict-based SGG score by 2.3\% over SFT ($0.672 \rightarrow 0.688$), mainly through better relation precision and relation F1.
In the harder generated-prompt setting, where previous-frame errors may propagate, the gain is larger: the Strict-based SGG score improves by 19\% ($0.306 \rightarrow 0.365$), outperforming all commercial, open-source, and SGG baselines.
Across PSG and PVSG, SceneGraphVLM keeps inference around one second per graph ($0.643$\,s and $1.068$\,s, respectively).

\begin{table*}[t]
\centering
\tiny
\renewcommand{\arraystretch}{1.08}
\setlength{\tabcolsep}{2.1pt}
\caption{
Full results on the PVSG dataset at IoU threshold $=50$.
We report generation time, failure rate, object metrics, relation metrics, and the final SGG score
under Strict and Soft evaluation. Results are grouped by temporal prompting protocol.
GT prompt and Generated prompt are treated as two independent evaluation sections.
Best overall results within each section are shown in bold; best results within each model group are underlined.
}
\label{tab:pvsg_iou50_label_modes_combined}

\resizebox{\textwidth}{!}{%
\begin{tabular}{@{}l c c c *{7}{cc}@{}}
\toprule
\multirow{2}{*}{\textbf{Model}}
& \multirow{2}{*}{\textbf{\makecell{Train\\split}}}
& \multirow{2}{*}{\textbf{\makecell{Gen. time\\ (s)}}}
& \multirow{2}{*}{\textbf{\makecell{Fail.\\rate (\%)}}}
& \multicolumn{2}{c}{\textbf{Obj Prec}}
& \multicolumn{2}{c}{\textbf{Obj Rec}}
& \multicolumn{2}{c}{\textbf{Obj F1}}
& \multicolumn{2}{c}{\textbf{Rel Prec}}
& \multicolumn{2}{c}{\textbf{Rel Rec}}
& \multicolumn{2}{c}{\textbf{Rel F1}}
& \multicolumn{2}{c}{\textbf{SGG Score}} \\
\cmidrule(lr){5-6}
\cmidrule(lr){7-8}
\cmidrule(lr){9-10}
\cmidrule(lr){11-12}
\cmidrule(lr){13-14}
\cmidrule(lr){15-16}
\cmidrule(l){17-18}
& & &
& \textbf{Strict} & \textbf{Soft}
& \textbf{Strict} & \textbf{Soft}
& \textbf{Strict} & \textbf{Soft}
& \textbf{Strict} & \textbf{Soft}
& \textbf{Strict} & \textbf{Soft}
& \textbf{Strict} & \textbf{Soft}
& \textbf{Strict} & \textbf{Soft} \\
\midrule

\multicolumn{18}{c}{\textbf{GT prompt}} \\
\midrule
\multicolumn{18}{@{}l@{}}{\textit{\textbf{Commercial M-LLMs}}} \\

GPT-5.4-nano
& -- & 2.678$\pm$0.414 & 2.23
& 0.144 & 0.202 & 0.129 & 0.183
& 0.129 & 0.181
& 0.019 & 0.022 & 0.021 & 0.022 & 0.019 & 0.020
& 0.074 & 0.101 \\

GPT-5-mini
& -- & \underline{2.403$\pm$0.408} & 0.330
& 0.367 & 0.429 & 0.312 & 0.363
& 0.327 & 0.380
& 0.163 & 0.166 & 0.145 & 0.148 & 0.147 & 0.150
& 0.237 & 0.265 \\

GPT-5.4-mini
& -- & 2.488$\pm$0.408 & 0.550
& 0.437 & 0.449 & 0.463 & 0.476
& 0.441 & 0.454
& 0.247 & 0.253 & 0.278 & 0.285 & 0.256 & 0.263
& 0.349 & 0.358 \\

Gemini-3-flash-preview
& -- & 3.924$\pm$0.281 & \underline{0.160}
& \underline{0.512} & \underline{0.521}
& \underline{0.517} & \underline{0.527}
& \underline{0.507} & \underline{0.516}
& \underline{0.331} & \underline{0.334}
& \underline{0.347} & \underline{0.351}
& \underline{0.332} & \underline{0.335}
& \underline{0.419} & \underline{0.426} \\

\midrule
\multicolumn{18}{@{}l@{}}{\textit{\textbf{Open-source M-LLMs}}} \\

InternVL2.5-1B
& -- & \underline{1.169$\pm$0.068} & 2.03
& 0.001 & 0.002 & 0.000 & 0.001
& 0.000 & 0.001
& 0.000 & 0.000 & 0.000 & 0.000 & 0.000 & 0.000
& 0.000 & 0.000 \\

SmolVLM-500M-Instruct
& -- & 2.781$\pm$0.621 & 91.62
& 0.000 & 0.003 & 0.000 & 0.001
& 0.000 & 0.001
& 0.000 & 0.000 & 0.000 & 0.000 & 0.000 & 0.000
& 0.000 & 0.001 \\

InternVL3-1B
& -- & 2.541$\pm$0.556 & 4.39
& 0.010 & 0.018 & 0.005 & 0.008
& 0.006 & 0.011
& 0.000 & 0.000 & 0.000 & 0.000 & 0.000 & 0.000
& 0.003 & 0.006 \\

DeepSeek-VL2-tiny-3B
& -- & 3.574$\pm$0.633 & 0.49
& 0.016 & 0.062 & 0.005 & 0.020
& 0.008 & 0.029
& 0.001 & 0.001 & 0.000 & 0.000 & 0.000 & 0.000
& 0.004 & 0.015 \\

Qwen3-VL-2B
& -- & 7.550$\pm$0.553 & 56.06
& 0.026 & 0.334 & 0.012 & 0.089
& 0.015 & 0.131
& 0.024 & 0.025 & 0.012 & 0.013 & 0.015 & 0.016
& 0.015 & 0.073 \\

Ovis2.5-2B-Instruct
& -- & 5.693$\pm$0.423 & 70.28
& 0.042 & 0.101 & 0.031 & 0.069
& 0.034 & 0.078
& 0.013 & 0.015 & 0.007 & 0.007 & 0.008 & 0.009
& 0.021 & 0.043 \\

InternVL3.5-1B
& -- & 2.798$\pm$0.677 & 12.83
& 0.077 & 0.123 & 0.055 & 0.084
& 0.061 & 0.095
& 0.002 & 0.003 & 0.003 & 0.004 & 0.002 & 0.003
& 0.032 & 0.049 \\

Qwen3-VL-4B
& -- & 7.677$\pm$0.625 & 21.86
& 0.073 & 0.422 & 0.032 & 0.146
& 0.042 & 0.199
& 0.040 & 0.043 & 0.018 & 0.020 & 0.022 & 0.024
& 0.032 & 0.112 \\

Qwen3.5-0.8B
& -- & 1.178$\pm$0.513 & 15.33
& 0.184 & 0.300 & 0.207 & 0.343
& 0.193 & 0.314
& 0.032 & 0.034 & 0.028 & 0.030 & 0.028 & 0.030
& 0.119 & 0.190 \\

Qwen3.5-2B
& -- & 1.834$\pm$0.529 & 10.08
& 0.224 & 0.381 & 0.227 & 0.370
& 0.210 & 0.349
& 0.079 & 0.085 & 0.121 & 0.133 & 0.084 & 0.091
& 0.138 & 0.203 \\

Qwen3.5-4B
& -- & 2.995$\pm$0.581 & 9.80
& \underline{0.419} & \underline{0.486}
& \underline{0.389} & \underline{0.424}
& \underline{0.396} & \underline{0.438}
& \underline{0.184} & \underline{0.184}
& \underline{0.177} & \underline{0.177}
& \underline{0.177} & \underline{0.177}
& \underline{0.286} & \underline{0.307} \\

\midrule
\multicolumn{18}{@{}l@{}}{\textit{\textbf{Fine-tuned non-final variants}}} \\

\makecell[l]{InternVL2.5-1B (SFT)}
& MaxInfo & \underline{1.169$\pm$0.068} & \underline{0.00}
& \underline{0.721} & \underline{0.723}
& 0.710 & 0.713
& 0.710 & 0.712
& 0.510 & 0.511 & 0.500 & 0.500 & 0.499 & 0.500
& 0.605 & 0.606 \\

\makecell[l]{InternVL2.5-1B (SFT)}
& BaseAnnot & \underline{1.169$\pm$0.068} & \underline{0.00}
& 0.708 & 0.709
& \underline{0.727} & \underline{0.728}
& \underline{0.715} & \underline{0.716}
& 0.504 & 0.505 & 0.506 & 0.507 & 0.496 & 0.497
& 0.605 & 0.607 \\

\makecell[l]{InternVL2.5-1B (SFT)}
& PSFR & \underline{1.169$\pm$0.068} & 0.03
& 0.714 & 0.716 & 0.709 & 0.711
& 0.706 & 0.707
& \underline{0.520} & \underline{0.521}
& \underline{0.512} & \underline{0.513}
& \underline{0.510} & \underline{0.511}
& \underline{0.608} & \underline{0.609} \\

\midrule

TRaSER
& -- & -- & \textbf{0.00}
& 0.118 & 0.265 & 0.117 & 0.261
& 0.117 & 0.263
& 0.037 & 0.038 & 0.096 & 0.101 & 0.047 & 0.049
& 0.082 & 0.156 \\

\makecell[l]{R1-SGG (SFT+RL)}
& -- & 10.965$\pm$0.657 & 0.66
& 0.349 & 0.570
& 0.298 & 0.496
& 0.313 & 0.515
& 0.043 & 0.050
& 0.279 & 0.324
& 0.069 & 0.081
& 0.191 & 0.298 \\

\makecell[l]{SceneGraphVLM (SFT)}
& PSFR & \textbf{1.068$\pm$0.055} & \textbf{0.00}
& 0.725 & 0.727
& 0.725 & 0.727
& 0.720 & 0.722
& 0.635 & 0.636
& \textbf{0.631} & 0.632
& 0.622 & 0.624
& 0.671 & 0.673 \\

\makecell[l]{SceneGraphVLM (SFT)}
& MaxInfo & \textbf{1.068$\pm$0.055} & \textbf{0.00}
& 0.744 & 0.745
& 0.731 & 0.732
& 0.732 & 0.733
& 0.630 & 0.632
& 0.614 & 0.615
& 0.612 & 0.613
& 0.672 & 0.673 \\

\makecell[l]{SceneGraphVLM (SFT)}
& BaseAnnot & \textbf{1.068$\pm$0.055} & \textbf{0.00}
& 0.739 & 0.741
& \textbf{0.733} & \textbf{0.736}
& 0.729 & 0.732
& 0.631 & 0.633
& 0.623 & \textbf{0.648}
& 0.614 & 0.615
& 0.672 & 0.673 \\

\makecell[l]{SceneGraphVLM (SFT+RL)}
& BaseAnnot & \textbf{1.068$\pm$0.055} & \textbf{0.00}
& \textbf{0.779} & \textbf{0.781}
& 0.731 & 0.733
& \textbf{0.748} & \textbf{0.750}
& \textbf{0.673} & \textbf{0.674}
& 0.614 & 0.615
& \textbf{0.628} & \textbf{0.629}
& \textbf{0.688} & \textbf{0.690} \\

\midrule
\multicolumn{18}{c}{\textbf{Generated prompt}} \\
\midrule
\multicolumn{18}{@{}l@{}}{\textit{\textbf{Commercial M-LLMs}}} \\

GPT-5.4-nano
& -- & 2.714$\pm$0.809 & 2.47
& 0.038 & 0.144 & 0.031 & 0.115
& 0.032 & 0.119
& 0.004 & 0.005 & 0.003 & 0.004 & 0.003 & 0.004
& 0.017 & 0.062 \\

GPT-5.4-mini
& -- & \underline{2.400$\pm$0.261} & \underline{0.77}
& 0.116 & 0.300 & 0.089 & 0.214
& 0.096 & 0.237
& 0.043 & 0.058 & 0.025 & 0.034 & 0.027 & 0.038
& 0.062 & 0.137 \\

Gemini-3-flash-preview
& -- & 4.245$\pm$0.949 & 3.65
& \underline{0.239} & \underline{0.419}
& \underline{0.252} & \underline{0.449}
& \underline{0.235} & \underline{0.415}
& \underline{0.083} & \underline{0.098}
& \underline{0.158} & \underline{0.183}
& \underline{0.099} & \underline{0.116}
& \underline{0.144} & \underline{0.207} \\

\midrule
\multicolumn{18}{@{}l@{}}{\textit{\textbf{Open-source M-LLMs}}} \\

SmolVLM-500M-Instruct
& -- & 2.781$\pm$0.621 & 96.40
& 0.000 & 0.000 & 0.000 & 0.000
& 0.000 & 0.000
& 0.000 & 0.000 & 0.000 & 0.000 & 0.000 & 0.000
& 0.000 & 0.000 \\

DeepSeek-VL2-tiny-3B
& -- & 3.574$\pm$0.633 & 4.92
& 0.002 & 0.022 & 0.001 & 0.019
& 0.002 & 0.020
& 0.000 & 0.000 & 0.000 & 0.000 & 0.000 & 0.000
& 0.001 & 0.010 \\

Qwen3-VL-2B
& -- & 7.550$\pm$0.553 & 97.50
& 0.004 & 0.011 & 0.003 & 0.008
& 0.004 & 0.009
& 0.001 & 0.001 & 0.001 & 0.001 & 0.001 & 0.001
& 0.002 & 0.005 \\

InternVL3-1B
& -- & 2.541$\pm$0.556 & 3.38
& 0.011 & 0.036 & 0.005 & 0.016
& 0.006 & 0.021
& 0.000 & 0.000 & 0.000 & 0.000 & 0.000 & 0.000
& 0.003 & 0.010 \\

InternVL2.5-1B
& -- & \underline{1.169$\pm$0.068} & 0.19
& 0.015 & 0.017 & 0.006 & 0.007
& 0.008 & 0.009
& 0.000 & 0.000 & 0.000 & 0.000 & 0.000 & 0.000
& 0.004 & 0.005 \\

Qwen3.5-0.8B
& -- & 1.178$\pm$0.513 & 44.69
& 0.026 & 0.138 & 0.014 & 0.044
& 0.015 & 0.056
& 0.001 & 0.001 & 0.001 & 0.001 & 0.001 & 0.001
& 0.008 & 0.028 \\

InternVL3.5-1B
& -- & 2.798$\pm$0.677 & 3.60
& 0.020 & 0.038 & 0.020 & 0.033
& 0.018 & 0.032
& 0.000 & 0.000 & 0.000 & 0.000 & 0.000 & 0.000
& 0.009 & 0.016 \\

Ovis2.5-2B-Instruct
& -- & 5.693$\pm$0.423 & 12.69
& 0.038 & 0.066 & 0.063 & 0.107
& 0.045 & 0.076
& 0.001 & 0.001 & 0.001 & 0.001 & 0.001 & 0.001
& 0.023 & 0.038 \\

Qwen3.5-2B
& -- & 1.834$\pm$0.529 & 8.13
& 0.078 & 0.185 & 0.080 & 0.191
& 0.072 & 0.171
& 0.009 & 0.011 & 0.007 & 0.009 & 0.007 & 0.008
& 0.039 & 0.090 \\

Qwen3.5-4B
& -- & 2.995$\pm$0.581 & 3.35
& 0.092 & 0.232
& \underline{0.125} & \underline{0.298}
& 0.097 & 0.236
& \underline{0.047} & 0.054
& \underline{0.084} & 0.092
& 0.048 & 0.054
& 0.073 & 0.145 \\

Qwen3-VL-4B
& -- & 7.677$\pm$0.625 & 1.92
& \underline{0.144} & \underline{0.405}
& 0.109 & 0.279
& \underline{0.118} & \underline{0.311}
& \underline{0.047} & \underline{0.059}
& 0.082 & \underline{0.099}
& \underline{0.049} & \underline{0.061}
& \underline{0.084} & \underline{0.186} \\

\midrule
\multicolumn{18}{@{}l@{}}{\textit{\textbf{Fine-tuned non-final variants}}} \\

\makecell[l]{InternVL2.5-1B $\mid$ SFT}
& PSFR & \underline{1.169$\pm$0.068} & 0.16
& 0.405 & 0.415 & 0.341 & 0.355
& 0.360 & 0.370
& 0.057 & 0.065 & 0.041 & 0.049 & 0.049 & 0.053
& 0.119 & 0.205 \\

\makecell[l]{InternVL2.5-1B $\mid$ SFT}
& MaxInfo & \underline{1.169$\pm$0.068} & \underline{0.03}
& 0.409 & 0.425 & 0.363 & 0.378
& 0.374 & 0.389
& 0.068 & 0.078 & 0.045 & 0.051 & 0.050 & 0.056
& 0.212 & 0.223 \\

\makecell[l]{InternVL2.5-1B $\mid$ SFT}
& BaseAnnot & \underline{1.169$\pm$0.068} & \underline{0.03}
& \underline{0.473} & \underline{0.487}
& \underline{0.388} & \underline{0.404}
& \underline{0.418} & \underline{0.420}
& \underline{0.122} & \underline{0.131}
& \underline{0.097} & \underline{0.107}
& \underline{0.099} & \underline{0.107}
& \underline{0.243} & \underline{0.254} \\

\midrule
\multicolumn{18}{@{}l@{}}{\textit{\textbf{Final models}}} \\

\makecell[l]{R1-SGG (SFT+RL)}
& -- & 10.965$\pm$0.657 & 9.56
& 0.102 & 0.235
& 0.091 & 0.215
& 0.093 & 0.215
& 0.016 & 0.020
& 0.097 & 0.122
& 0.025 & 0.032
& 0.059 & 0.124 \\

\makecell[l]{SceneGraphVLM $\mid$ SFT}
& PSFR & \textbf{1.068$\pm$0.055} & \textbf{0.00}
& 0.405 & 0.420 & 0.377 & 0.395
& 0.375 & 0.391
& 0.161 & 0.175 & 0.125 & 0.137 & 0.133 & 0.145
& 0.254 & 0.268 \\

\makecell[l]{SceneGraphVLM $\mid$ SFT}
& MaxInfo & \textbf{1.068$\pm$0.055} & \textbf{0.00}
& 0.486 & 0.515 & 0.373 & 0.395
& 0.411 & 0.436
& 0.159 & 0.180 & 0.112 & 0.133 & 0.122 & 0.140
& 0.266 & 0.288 \\

\makecell[l]{SceneGraphVLM $\mid$ SFT}
& BaseAnnot & \textbf{1.068$\pm$0.055} & \textbf{0.00}
& 0.508 & 0.530 & 0.423 & 0.440
& 0.450 & 0.469
& 0.194 & 0.217 & 0.158 & 0.186 & 0.163 & 0.185
& 0.306 & 0.327 \\

\makecell[l]{SceneGraphVLM $\mid$ SFT+RL}
& BaseAnnot & \textbf{1.068$\pm$0.055} & \textbf{0.00}
& \textbf{0.574} & \textbf{0.599}
& \textbf{0.469} & \textbf{0.491}
& \textbf{0.504} & \textbf{0.528}
& \textbf{0.295} & \textbf{0.319}
& \textbf{0.202} & \textbf{0.222}
& \textbf{0.227} & \textbf{0.247}
& \textbf{0.365} & \textbf{0.388} \\

\bottomrule
\end{tabular}%
}
\end{table*}

\begin{figure*}[!h]
    \centering
    \includegraphics[width=\textwidth]{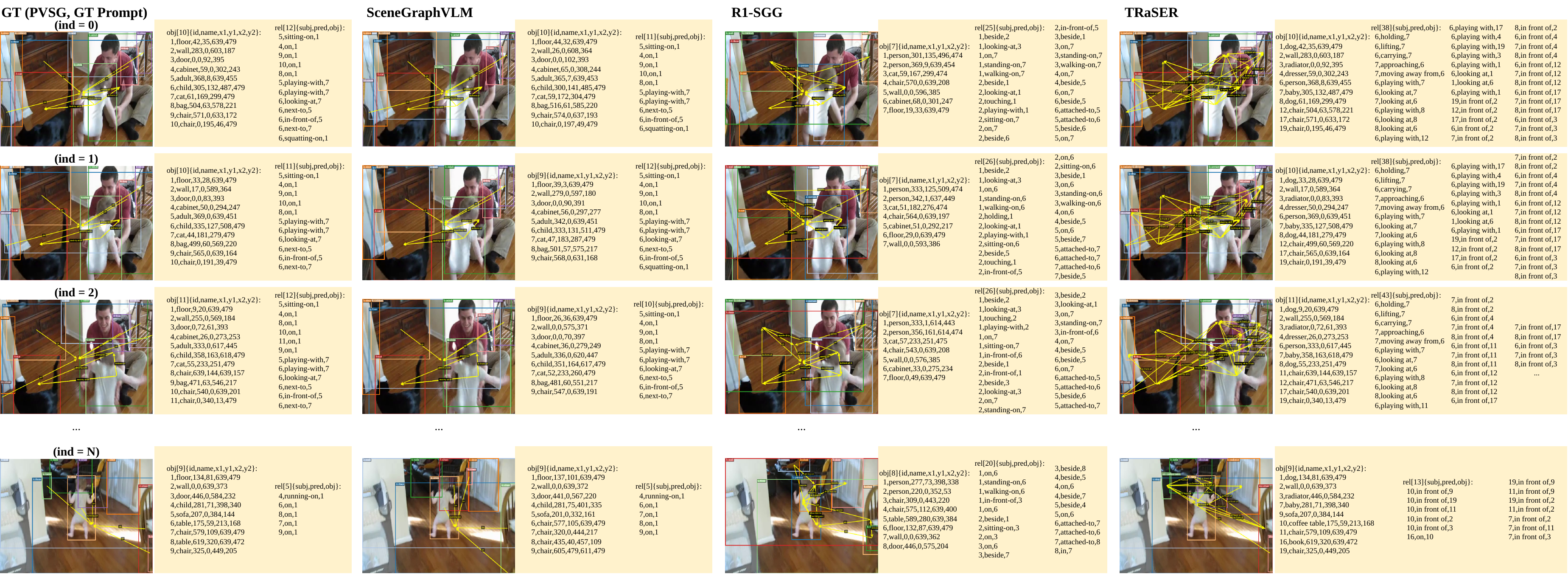}
    \caption{
    Qualitative comparison on PVSG in the GT-prompt setting.
    The previous-frame graph is provided from ground truth, so temporal context is clean and does not introduce accumulated errors.
    SceneGraphVLM produces compact and temporally consistent scene graphs that preserve the main objects and visually supported relations across adjacent frames.
    Compared with more recall-oriented baselines such as R1-SGG, our method avoids relation over-generation and maintains stricter, better grounded relations instead of adding many weakly supported edges.
    }
    \label{fig:pvsg_gt_qualitative}
\end{figure*}

\begin{figure*}[!h]
    \centering
    \includegraphics[width=\textwidth]{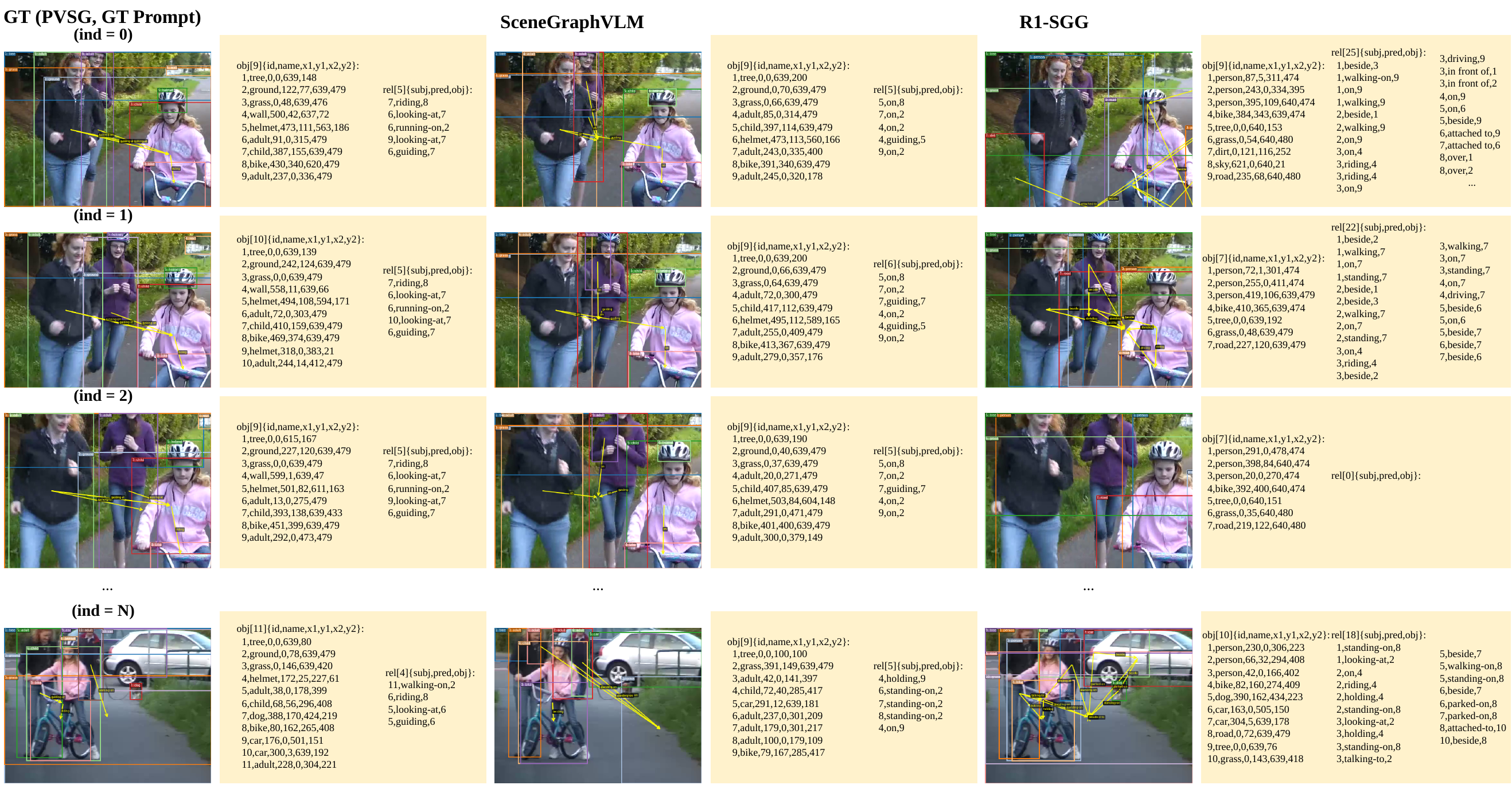}
    \caption{
    Qualitative comparison on PVSG in the Generated-prompt setting.
    Here the model receives its own previous prediction as temporal context, making the task substantially harder because object and relation errors may propagate over time.
    SceneGraphVLM remains more robust in this setting: it better preserves temporal consistency, suppresses spurious relations induced by noisy context, and avoids cascading relation hallucinations.
    Compared with denser baselines, our model produces a more conservative but cleaner graph, prioritizing grounded object--relation structure over raw relation count.
    }
    \label{fig:pvsg_gen_qualitative}
\end{figure*}

\textbf{Qualitative PVSG comparison.}
Figures~\ref{fig:pvsg_gt_qualitative} and~\ref{fig:pvsg_gen_qualitative} provide qualitative comparisons on PVSG under the GT-prompt and Generated-prompt settings, respectively.
In the GT-prompt setting, the previous-frame graph is provided from ground truth, so temporal context is clean and does not introduce accumulated errors.
In this regime, SceneGraphVLM produces compact and temporally consistent scene graphs that preserve the main objects and visually supported relations across adjacent frames.
Compared with more recall-oriented baselines such as R1-SGG, our method avoids relation over-generation and maintains stricter, better grounded relations instead of adding many weakly supported edges.
The main remaining errors are occasional missed secondary relations or small objects, but the predicted graph stays structurally valid and closely aligned with the visible scene.

In the Generated-prompt setting, the model receives its own previous prediction as temporal context, making the task substantially harder because object and relation errors may propagate over time.
SceneGraphVLM remains more robust in this setting: it better preserves temporal consistency, suppresses spurious relations induced by noisy context, and avoids cascading relation hallucinations.
Compared with denser baselines, our model produces a more conservative but cleaner graph, prioritizing grounded object--relation structure over raw relation count.
Its typical failure mode is under-generation when the previous predicted graph is incomplete, yet the final output remains more stable, precise, and practical for downstream use.

\textbf{Action Genome results.}
Table~\ref{tab:ag_sgdet_combined} reports results on Action Genome under the SGDET with-constraint protocol.
The upper part provides the standard recall-only comparison with prior methods, while the lower part extends the comparison with precision and F1 for OED and SceneGraphVLM.
Classical SGG methods, especially OED, achieve higher Recall@K because they retrieve ranked triplets from a large candidate set.
In contrast, SceneGraphVLM generates compact relation sets and substantially improves precision and F1, reaching $P@50=38.74$ and $F1@50=30.75$, compared with $P@50=8.02$ and $F1@50=13.37$ for OED.
Its metrics nearly saturate after $K=10$, since the model generates only a compact set of relation triplets per frame.

\begin{table}[!h]
\centering
\scriptsize
\renewcommand{\arraystretch}{0.92}
\setlength{\tabcolsep}{3pt}
\caption{
Comparison on Action Genome under the SGDET \textbf{With Constraint} setting.
The upper block reports standard Recall@K against prior methods.
The lower block reports extended Precision@K, Recall@K, and F1@K for OED and SceneGraphVLM.
Best values are highlighted in bold within each block.
}
\label{tab:ag_sgdet_combined}

\begin{tabular*}{\textwidth}{@{\extracolsep{\fill}}lccc lccc@{}}
\toprule
\multicolumn{8}{c}{\textbf{Recall-only comparison}} \\
\midrule
\textbf{Method} & \textbf{R@10} & \textbf{R@20} & \textbf{R@50}
&
\textbf{Method} & \textbf{R@10} & \textbf{R@20} & \textbf{R@50} \\
\midrule
VRD~\cite{lu2016visual}              & 19.2 & 24.5 & 26.0
& STTran~\cite{cong2021spatial}      & 25.2 & 34.1 & 37.0 \\
MSDN~\cite{li2017scene}              & 24.1 & 32.4 & 34.5
& APT~\cite{li2022dynamic}           & 26.3 & 36.1 & 38.3 \\
M-FREQ~\cite{zellers2018neural}      & 23.7 & 31.4 & 33.3
& STTran-TPI~\cite{wang2022dynamic}  & 26.2 & 34.6 & 37.4 \\
VCTree~\cite{tang2019learning}       & 24.4 & 32.6 & 34.7
& TR$^2$~\cite{wang2023cross}        & 26.8 & 35.5 & 38.3 \\
RelDN~\cite{zhang2019graphical}      & 24.5 & 32.8 & 34.9
& TEMPURA~\cite{nag2023unbiased}     & 28.1 & 33.4 & 34.9 \\
GPS-Net~\cite{lin2020gps}            & 24.7 & 33.1 & 35.1
& DSG-DETR~\cite{feng2023exploiting} & 30.3 & 34.8 & 36.1 \\
TRACE~\cite{teng2021target}          & 13.9 & 14.5 & 14.5
& OED~\cite{wang2024oed}             & \textbf{33.5} & \textbf{40.9} & \textbf{48.9} \\
RelTR~\cite{cong2023reltr}           & 19.7 & 23.4 & 25.9
& SceneGraphVLM (SFT+RL)             & 27.76 & 27.84 & 27.84 \\
\bottomrule
\end{tabular*}

\vspace{0.75em}

\begin{tabular*}{\textwidth}{@{\extracolsep{\fill}}lccccccccc@{}}
\toprule
\multicolumn{10}{c}{\textbf{Extended precision--recall--F1 comparison}} \\
\midrule
\multirow{2}{*}{\textbf{Method}}
& \multicolumn{3}{c}{\textbf{@10}}
& \multicolumn{3}{c}{\textbf{@20}}
& \multicolumn{3}{c}{\textbf{@50}} \\
\cmidrule(lr){2-4}
\cmidrule(lr){5-7}
\cmidrule(l){8-10}
& \textbf{P} & \textbf{R} & \textbf{F1}
& \textbf{P} & \textbf{R} & \textbf{F1}
& \textbf{P} & \textbf{R} & \textbf{F1} \\
\midrule
OED~\cite{wang2024oed}
& 26.65 & \textbf{33.52} & 27.81
& 16.65 & \textbf{41.01} & 22.46
& 8.02  & \textbf{48.99} & 13.37 \\

SceneGraphVLM (SFT+RL)
& \textbf{38.73} & 27.76 & \textbf{30.71}
& \textbf{38.74} & 27.84 & \textbf{30.75}
& \textbf{38.74} & 27.84 & \textbf{30.75} \\
\bottomrule
\end{tabular*}
\end{table}

\subsection{Ablation Studies}
\label{sec:ablations}

\textbf{Effect of reinforcement learning.}
As shown in Table~\ref{tab:pvsg_iou50_label_modes_combined}, RL consistently improves SceneGraphVLM over the SFT checkpoint on PVSG.
Under the GT-prompt setting, the Soft SGG score increases from $0.673$ to $0.690$, corresponding to a relative improvement of $2.5\%$.
Under the generated previous-graph setting, where temporal errors can accumulate, the gain is larger: the Soft SGG score improves from $0.327$ to $0.388$, corresponding to a relative improvement of $18.7\%$.
The improvements are especially visible in relation precision and relation F1, which supports the role of hallucination-aware rewards in suppressing unsupported relations while preserving useful graph coverage.

\textbf{Reward ablation.}
Table~\ref{tab:reward_ablation_pvsg_generated} ablates reward groups in the PVSG generated-prompt setting.
Base graph rewards improve recall-oriented metrics over SFT, while adding relation-balance terms increases relation precision and F1.
The full hallucination-aware reward gives the best SGG score, showing that explicit penalties are important for compact and reliable generated-graph inference.

\begin{table*}[t]
\centering
\tiny
\renewcommand{\arraystretch}{1.08}
\setlength{\tabcolsep}{2.1pt}
\caption{
Reward ablation on PVSG in the generated previous-graph setting at IoU threshold $50$.
All RL variants are initialized from the same SFT checkpoint.
``Base'' denotes format, object, and recall-oriented graph rewards;
``Bal.'' denotes relation precision and F1 rewards;
``Hall.'' denotes object and relation hallucination penalties.
}
\label{tab:reward_ablation_pvsg_generated}
\resizebox{\textwidth}{!}{%
\begin{tabular}{@{}l c c c c *{7}{cc}@{}}
\toprule
\multirow{2}{*}{\textbf{Setting}}
& \multicolumn{3}{c}{\textbf{Reward groups}}
& \multirow{2}{*}{\textbf{\makecell{Fail.\\rate (\%)}}}
& \multicolumn{2}{c}{\textbf{Obj Prec}}
& \multicolumn{2}{c}{\textbf{Obj Rec}}
& \multicolumn{2}{c}{\textbf{Obj F1}}
& \multicolumn{2}{c}{\textbf{Rel Prec}}
& \multicolumn{2}{c}{\textbf{Rel Rec}}
& \multicolumn{2}{c}{\textbf{Rel F1}}
& \multicolumn{2}{c}{\textbf{SGG Score}} \\
\cmidrule(lr){2-4}
\cmidrule(lr){6-7}
\cmidrule(lr){8-9}
\cmidrule(lr){10-11}
\cmidrule(lr){12-13}
\cmidrule(lr){14-15}
\cmidrule(lr){16-17}
\cmidrule(l){18-19}
& \textbf{Base} & \textbf{Bal.} & \textbf{Hall.}
& 
& \textbf{Strict} & \textbf{Soft}
& \textbf{Strict} & \textbf{Soft}
& \textbf{Strict} & \textbf{Soft}
& \textbf{Strict} & \textbf{Soft}
& \textbf{Strict} & \textbf{Soft}
& \textbf{Strict} & \textbf{Soft}
& \textbf{Strict} & \textbf{Soft} \\
\midrule

SFT
& -- & -- & --
& 0.00
& 0.508 & 0.530
& 0.423 & 0.440
& 0.450 & 0.469
& 0.194 & 0.217
& 0.158 & 0.186
& 0.163 & 0.185
& 0.306 & 0.327 \\

Base RL
& $\checkmark$ & -- & --
& 0.00
& 0.522 & 0.546
& 0.461 & 0.483
& 0.478 & 0.500
& 0.183 & 0.195
& \textbf{0.208} & 0.219
& 0.180 & 0.190
& 0.329 & 0.345 \\

Balanced RL
& $\checkmark$ & $\checkmark$ & --
& 0.00
& 0.533 & 0.559
& 0.454 & 0.476
& 0.478 & 0.501
& 0.224 & 0.248
& 0.192 & 0.213
& 0.192 & 0.213
& 0.335 & 0.357 \\

Full RL
& $\checkmark$ & $\checkmark$ & $\checkmark$
& 0.00
& \textbf{0.574} & \textbf{0.599}
& \textbf{0.469} & \textbf{0.491}
& \textbf{0.504} & \textbf{0.528}
& \textbf{0.295} & \textbf{0.319}
& 0.202 & \textbf{0.222}
& \textbf{0.227} & \textbf{0.247}
& \textbf{0.365} & \textbf{0.388} \\

\bottomrule
\end{tabular}%
}
\end{table*}

\textbf{Frame thinning.}
PVSG contains many near-duplicate adjacent frames, which makes full-frame training redundant and prone to overfitting to local continuity.
Frame thinning improves both efficiency and graph quality.
As shown in Table~\ref{tab:pvsg_iou50_label_modes_combined}, PSFR and annotation-based thinning substantially outperform full-frame training under GT prompt.
Under generated prompt, annotation-based thinning gives the most robust results, suggesting that removing redundant temporal samples improves generalization when previous-frame context is model-generated rather than ground truth.

\section{Limitations}
\label{sec:limitations}

Our work has several limitations. First, SceneGraphVLM is trained and evaluated mainly on PSG, PVSG, and Action Genome, so it may inherit dataset-specific biases, annotation noise, and long-tail category imbalance. Second, the current setup mostly follows a closed-vocabulary formulation, which may limit generalization to unseen object categories and relation types. Third, in the video setting, using a generated previous-frame graph as temporal context can lead to error accumulation across frames, especially when the initial prediction is inaccurate. Fourth, although the TOON representation and vLLM inference substantially reduce generation latency, practical near-real-time deployment still requires appropriate GPU resources. Fifth, for PVSG we evaluate on a PSFR-selected test subset to reduce temporal redundancy; although this improves evaluation efficiency, it is not identical to evaluating on every annotated PVSG frame. Finally, our current video formulation focuses on frame-level object and relation prediction with short-term temporal context; extending the method to more persistent long-horizon temporal reasoning is an important direction for future work.

\section{Conclusion}
\label{sec:conclusion}

We presented SceneGraphVLM, a compact VLM-based method for image and video scene graph generation.
It combines the token-efficient TOON representation with supervised fine-tuning and hallucination-aware GRPO rewards for grounded, precise graph prediction.
Across PSG, PVSG, and Action Genome, SceneGraphVLM improves the quality--speed trade-off of generative SGG, outperforming zero-shot VLM baselines while remaining efficient with vLLM decoding.
These results highlight compact graph representations and precision-aware training as key ingredients for practical VLM-based scene graph generation in downstream and near-real-time video settings.

\bibliographystyle{plainnat}
\bibliography{references}

\end{document}